\documentclass{article}

\usepackage[preprint, 
            nonatbib,
            ]{neurips_2023}
\usepackage{graphicx}
\usepackage{subcaption}
\usepackage[dvipsnames,svgnames,x11names]{xcolor}
\usepackage[utf8]{inputenc} 
\usepackage[T1]{fontenc}    
\usepackage{hyperref}       
\usepackage{url}            
\usepackage{booktabs}       
\usepackage{amsfonts}       
\usepackage{nicefrac}       
\usepackage{microtype}      
\usepackage{multicol}
\usepackage{multirow}
\usepackage{adjustbox}
\usepackage{wrapfig}
\usepackage{changes}
\usepackage{xspace}
\usepackage{float}
\usepackage{cleveref}

\definechangesauthor[color=BrickRed]{SS}

\newenvironment{packed_enumerate}
{\begin{enumerate}
    \setlength{\itemsep}{1pt}
    \setlength{\parskip}{0pt}
    \setlength{\parsep}{0pt}
}{\end{enumerate}}

\newcommand{\filluptopage}[1]{%
  \clearpage
  \loop\ifnum\value{page}<#1\relax
    \null\clearpage
  \repeat
  \loop\ifnum\value{page}=#1\relax
    \null\clearpage
  \repeat
}

\makeatletter
\def\blfootnote{\xdef\@thefnmark{}\@footnotetext}
\makeatother

\newcommand{\acro}{HyP-NeRF\xspace}
\title{\acro: Learning Improved \underline{NeRF} \underline{P}riors using a \underline{Hy}perNetwork}

\author{%
  Bipasha Sen\thanks{Equal authors (order decided by a coin flip)} \\
  MIT CSAIL\\
  \texttt{bise@mit.edu} \\
  \And
  Gaurav Singh$^*$ \\
  IIIT, Hyderabad \\
    \texttt{gaurav.si@research.iiit.ac.in} \\
  \And
  Aditya Agarwal$^*$ \\
  MIT CSAIL\\
  \texttt{adityaag@mit.edu} \\
  \And 
  Rohith Agaram \\
  IIIT, Hyderabad \\
  \texttt{rohith.agaram@research.iiit.ac.in} \\
  \And
  K Madhava Krishna \\
  IIIT, Hyderabad \\
  \texttt{mkrishna@iiit.ac.in} \\
  \And
  Srinath Sridhar \\
  Brown University \\
  \texttt{srinath@brown.edu} \\
}

\begin{document}

\maketitle

\begin{figure}[th!]
    \centering 
    \vspace{-25px}
    \includegraphics[width=\linewidth]{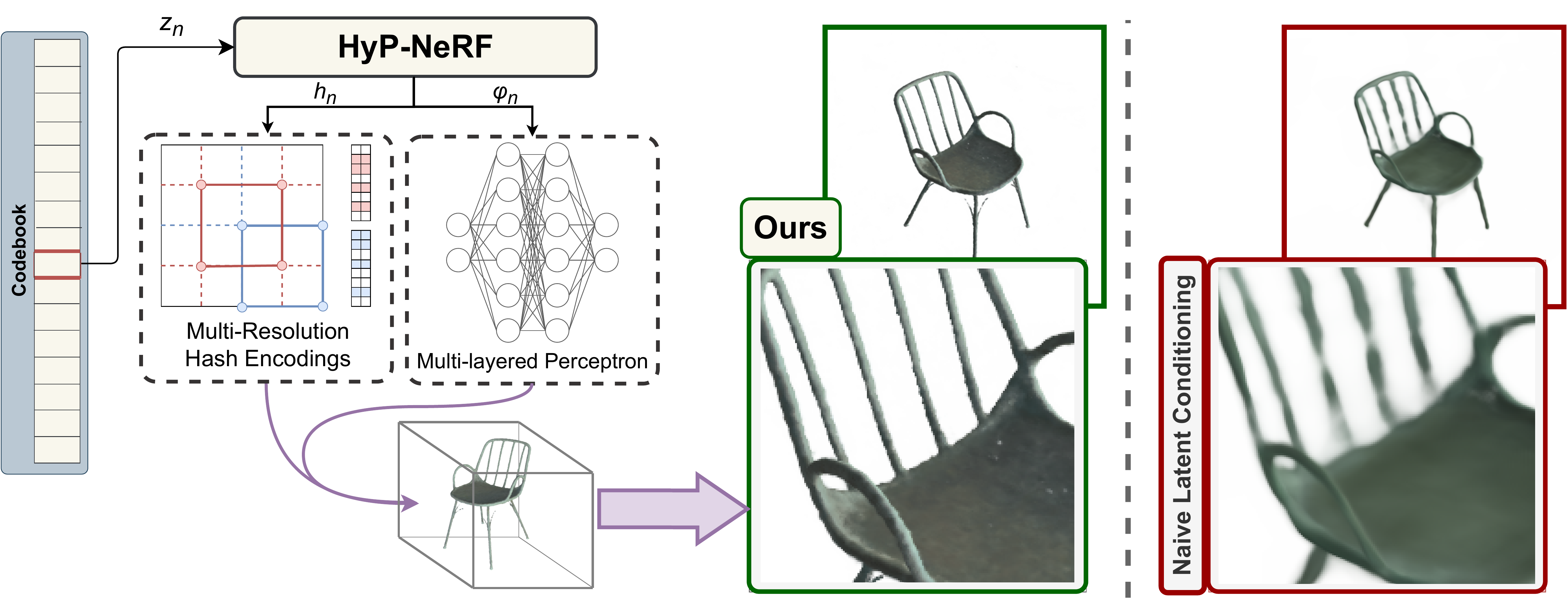}
    \vspace{-1.4em}
    \caption{We propose \acro, a latent conditioning method that learns improved quality NeRF priors using a hypernetwork to generate instance-specific multi-resolution hash encodings along with neural network weights. The figure showcases the fine details preserved in the NeRF generated by \acro (green box) as opposed to the NeRF generated by naive conditioning (red box) in which, a hypernetwork predicts only the neural weights while relying on the standard positional encodings.}
    \label{fig:teaser}
\end{figure}

\begin{abstract}
Neural Radiance Fields (NeRF) have become an increasingly popular representation to capture high-quality appearance and shape of scenes and objects.
However, learning generalizable NeRF priors over categories of scenes or objects has been challenging due to the high dimensionality of network weight space.
To address the limitations of existing work on generalization, multi-view consistency and to improve quality, we propose \acro, a latent conditioning method for learning generalizable category-level NeRF priors using hypernetworks.
Rather than using hypernetworks to estimate only the weights of a NeRF, we estimate both the weights and the multi-resolution hash encodings~\cite{mueller2022instant} resulting in significant quality gains.
To improve quality even further, we incorporate a denoise and finetune strategy that denoises images rendered from NeRFs estimated by the hypernetwork and finetunes it while retaining multiview consistency.
These improvements enable us to use \acro as a generalizable prior for multiple downstream tasks including NeRF reconstruction from single-view or cluttered scenes, and text-to-NeRF.
We provide qualitative comparisons and evaluate \acro on three tasks: generalization, compression, and retrieval, demonstrating our state-of-the-art results. 
\footnote{Project page: \textcolor{blue}{\href{https://hyp-nerf.github.io}{hyp-nerf.github.io}}}

\end{abstract}

\section{Introduction}

Neural fields, also known as implicit neural representations (INRs), are neural networks that learn a continuous representation of physical quantities such as shape or radiance at any given space-time coordinate~\cite{xie2022_neuralfields}.
Recent developments in neural fields have enabled significant advances in applications such as 3D shape generation~\cite{neuralfield3Dshapegen}, novel view synthesis~\cite{nerf,mipnerf}, 3D reconstruction~\cite{volsdf,neus,unisurf,neuralfield3Dreconstruction}, and robotics~\cite{ndfs, rndfs}.
In particular, we are interested in Neural Radiance Fields (NeRF) that learn the parameters of a neural network $f_\phi(\mathbf{x}, \theta) = \{\sigma, c\}$, where $\mathbf{x}$ and $\theta$ are the location and viewing direction of a 3D point, respectively, and $\sigma$ and $c$ denote the density and color estimated by $f_\phi$ at that point.
Once fully trained, $f_\phi$ can be used to render novel views of the 3D scene.

Despite their ability to model high-quality appearance, NeRFs cannot easily generalize to scenes or objects not seen during training thus limiting their broader application.
Typically, achieving generalization involves learning a prior over a data source 
such as image, video, or point cloud distributions~\cite{stylegan, stylegan2-ada, stylegan-v, digan, treegan, scarp}, possibly belonging to a category of objects~\cite{nocs, sajnani2022_condor}.
However, NeRFs are continuous volumetric functions parameterized by tens of millions of parameters making it challenging to learn generalizable priors.
Previous works try to address this challenge by relying on 2D image-based priors, 3D priors in voxelized space, or by using latent conditioning.

Image-based priors re-use the information learned by 2D convolutional networks~\cite{pixelnerf, autorf} but may lack 3D knowledge resulting in representations that are not always multiview consistent. 
Methods that learn 3D priors in voxelized space~\cite{diffrf} suffer from high compute costs and inherently lower quality due to voxelization limitations.
Latent conditioning methods~\cite{jang2021codenerf, lolnerf} learn a joint network $f(\mathrm{x}, \theta, z)$ where $z$ is the conditioning vector for a given object instance.
These methods retain the advantages of native NeRF representations such as instance-level 3D and multiview consistency, but have limited capacity to model a diverse set of objects at high visual and geometric quality.
InstantNGP~\cite{mueller2022instant} provides a way to improve quality and speed using \emph{instance-specific} multi-resolution hash encodings (MRHE), however, this is limited to single instances.

\begin{wrapfigure}[23]{R}{6.8cm}
\vspace{-.8em}
    \centering 
    \includegraphics[width=\linewidth]{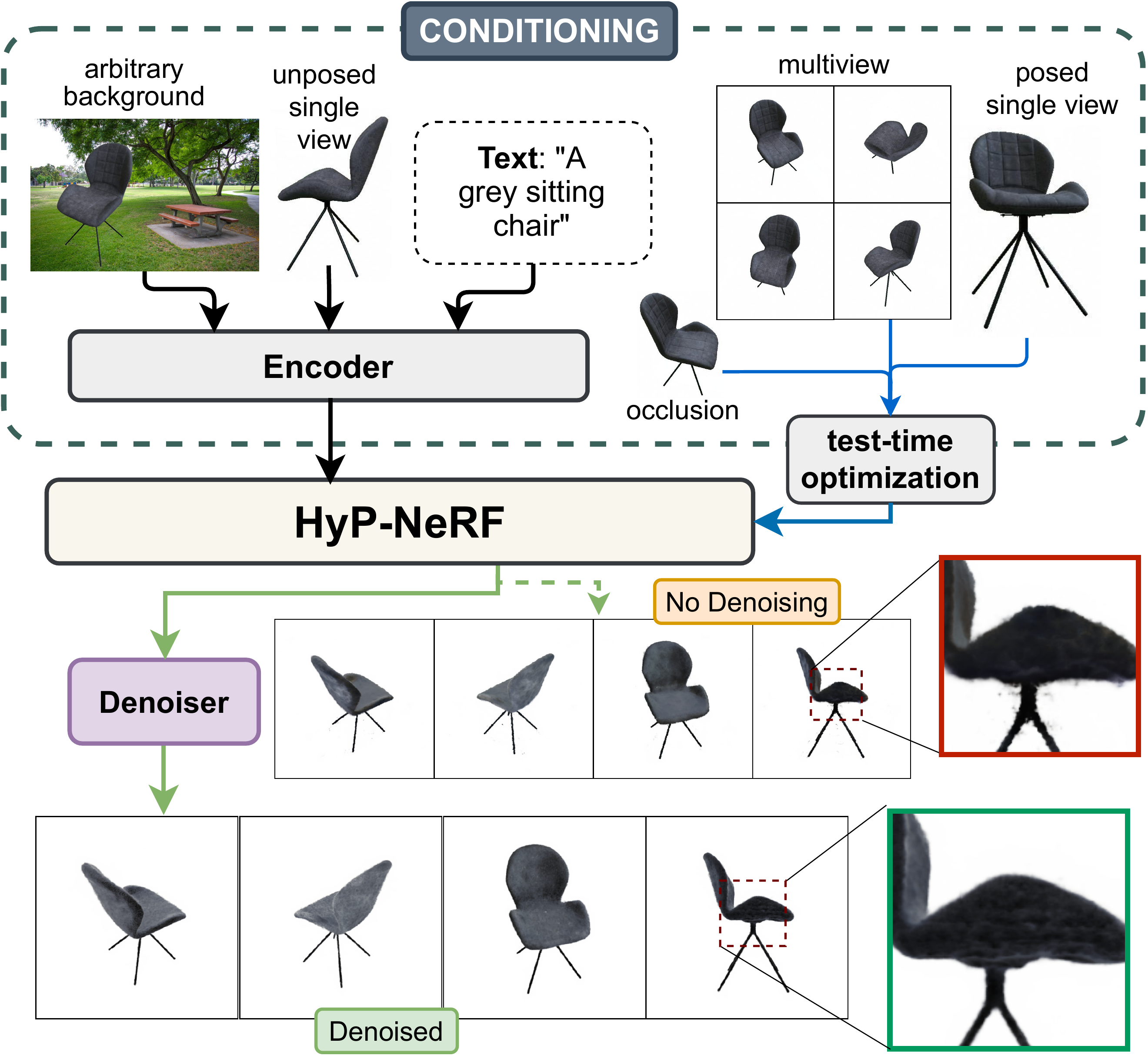}
    \vspace{-1.2em}
    \caption{
    \small Once trained, \acro acts as a prior to support multiple downstream applications, including NeRF reconstruction from single or multi-view images and cluttered scene images, and text-to-NeRF.
    We further improve quality using our denoising network.
    }
    \label{fig:applications}
\end{wrapfigure}
We propose \acro, a latent conditioning method for learning improved quality generalizable \textbf{category-level \underline{NeRF} \underline{p}riors} using \underline{hy}pernetworks~\cite{hypernetwork} (see \Cref{fig:teaser}).
We take inspiration from methods that use meta-learning to learn generalizable representations~\cite{lfns,sen2022inrv} while retaining the quality of instance-specific methods~\cite{mueller2022instant}.
Our hypernetwork is trained to generate the parameters--both the multi-resolution \textbf{hash encodings (MRHE) and weights}--of a NeRF model of a given category conditioned on an instance code $z_n$.
For each instance code $z_n$ in the learned codebook, \acro estimates $h_n$ denoting the instance-specific MRHE along with $\phi_n$ indicating the weights of an MLP.
Our key insight is that estimating both the MRHEs and the weights results in a significant improvement in quality.
To improve the quality even further, we denoise rendered views~\cite{vqvae2} from the estimated NeRF model, and finetune the NeRF with the denoised images to enforce multiview consistency.
As shown in \Cref{fig:applications} and the experiments section, this denoising and finetuning step significantly improves quality and fine details while retaining the original shape and appearance properties.

Once \acro is trained, it can be used as a NeRF prior in a variety of different applications such as NeRF reconstruction from a single view posed or unposed images, single pass text-to-NeRF, or even the ability to reconstruct real-world objects in cluttered scene images (see \Cref{fig:applications}).
We show qualitative results on applications and quantitatively evaluate \acro's performance and suitability as a NeRF prior on the ABO dataset~\cite{abo} across three tasks: generalization, compression, and retrieval.
To sum up our contributions:
\begin{packed_enumerate}
    \item We introduce \acro, a method for learning improved quality NeRF priors using a hypernetwork that estimates \textit{instance-specific} hash encodings and MLP weights of a NeRF. 
    \item We propose a denoise and finetune strategy to further improve the quality while preserving the multiview consistency of the generated NeRF.
    \item We demonstrate how our NeRF priors can be used in multiple downstream tasks including single-view NeRF reconstruction, text-to-NeRF, and reconstruction from cluttered scenes.
\end{packed_enumerate}

\vspace{-10px}
\section{Related Work}
\textbf{Neural Radiance Fields}~\cite{nerf} (NeRFs) are neural networks that capture a specific 3D scene or object given sufficient views from known poses.
Numerous follow-up work (see \cite{tewari2022advances,xie2022_neuralfields} for a more comprehensive review) has investigated improving quality and speed, relaxing assumptions, and building generalizable priors.
Strategies for improving quality or speed include better sampling~\cite{mipnerf}, supporting unbounded scenes~\cite{barron2022mipnerf360}, extensions to larger scenes~\cite{xiangli2021citynerf,tancik2022block}, using hybrid representations~\cite{mueller2022instant,yu2021plenoctrees}, using learned initializations~\cite{learnedinit, fwd, sajjadi2022scene}, or discarding neural networks completely~\cite{yu2021plenoxels,sun2022direct}.
Other work relaxes assumption of known poses~\cite{nerfminus, gnerf, barf, nerfpose, garf, sparsepose}, or reduce the number of views~\cite{pixelnerf, lolnerf, pix2nerf, guo2022fast, autorf, depthpriorsnerf, xu2022sinnerf, lin2023visionnerf,niemeyer2022regnerf}.
Specifically, PixelNeRF~\cite{pixelnerf} uses convolution-based image features to learn priors enabling NeRF reconstruction from as few as a single image.
VisionNeRF~\cite{lin2023visionnerf} extends PixelNeRF by augmenting the 2D priors with 3D representations learned using a transformer.
Unlike these methods, we depend purely on priors learned by meta-learning, specifically by hypernetworks~\cite{hypernetwork}.
AutoRF~\cite{autorf} and LolNeRF~\cite{lolnerf} are related works that assume only a single view for each instance at the training time.
FWD~\cite{fwd} optimizes NeRFs from sparse views in real-time and SRT~\cite{sajjadi2022scene} aims to generate NeRFs in a single forward pass.
These methods produce NeRFs of lower quality and are not designed to be used as priors for various downstream tasks.
In contrast, our focus is to generate high-quality multiview consistent NeRFs that capture fine shapes and textures details.
\acro can be used as a category-level prior for multiple downstream tasks including NeRF reconstruction from one or more posed or unposed images, text-to-NeRF (similar to \cite{dreamfusion,jain2022zero}), or reconstruction from cluttered scene images.
Additionally, \acro can estimate the NeRFs in a single forward pass with only a few iterations needed to improve the quality. 
Concurrent to our work, NerfDiff \cite{gu2023nerfdiff} and SSDNeRF \cite{ssdnerf} achieve high quality novel view synthesis by using diffusion models.

\textbf{Learning 3D Priors}.
To learn category-level priors, methods like CodeNeRF~\cite{jang2021codenerf} and LolNeRF~\cite{lolnerf} use a conditional NeRF on instance vectors $z$ given as $f(\mathrm{x}, \theta, z)$, where different $z$s result in different NeRFs.
PixelNeRF~\cite{pixelnerf} depends on 2D priors learned by 2D convolutional networks which could result in multi-view inconsistency.
DiffRf~\cite{diffrf} uses diffusion to learn a prior over voxelized radiance field. Like us, DiffRF can generate radiance fields from queries like text or images. However, it cannot be directly used for downstream tasks easily.

Our approach closely follows the line of work that aims to learn a prior over a 3D data distribution like signed distance fields~\cite{deepsdf}, light field~\cite{lfns}, and videos~\cite{sen2022inrv}.
We use meta-learning, specifically hypernetworks~\cite{hypernetwork}, to learn a prior over the MRHEs and MLP weights of a fixed NeRF architecture.
LearnedInit~\cite{learnedinit}, also employs standard meta-learning algorithms for getting a good initialization of the NeRF parameters. However, unlike us, they do not use a hypernetwork, and use the meta-learning algorithms only for initializing a NeRF, which is further finetuned on the multiview images.
Methods like GRAF~\cite{graf}, $\pi$-GAN~\cite{pigan}, CIPS-3D~\cite{zhou2021CIPS3D}, EG3D~\cite{eg3d}, and Pix2NeRF~\cite{pix2nerf} use adversarial training setups with 2D discriminators resulting in 3D and multiview inconsistency.
\cite{dreamfusion, clipnerf, instructnerf2nerf} tightly couple text and NeRF priors to generate and edit NeRFs based on text inputs.
We, on the other hand, train a 3D prior on NeRFs and separately train a mapping network that maps text to \acro's prior, decoupling the two.




\section{\acro: Learning Improved NeRF prior using a Hypernetwork}
\label{sec:method}
Our goal is to learn a generalizable NeRF prior for a category of objects while maintaining visual and geometric quality, and multiview consistency.
We also want to demonstrate how this prior can be used to enable downstream applications in single/few-image NeRF generation, text-to-NeRF, and reconstruction of real-world objects in cluttered scenes.

\textbf{Background}.
We first provide a brief summary of hypernetworks and multi-resolution hash encodings that form the basis of \acro.
Hypernetworks are neural networks that were introduced as a meta-network to predict the weights for a second neural network. They have been widely used for diverse tasks, starting from representation learning for continuous signals~\cite{lfns, siren, inr-gan, sen2022inrv}, compression~\cite{hncompress1, hncompress2}, few-shot learning~\cite{hnfewshot1, hnfewshot2}, continual learning~\cite{continualhypernet}. Our key insight is to use hypernetworks to generate both the network weights and instance-specific MRHEs.

\textbf{Neural Radiance Fields }(NeRF)~\cite{nerf, mipnerf} learn the parameters of a neural network $f_\phi(\mathbf{x}, \theta) = \{\sigma, c\}$, where $\mathbf{x}$ and $\theta$ are the location and viewing direction of a 3D point, respectively, and $\sigma$ and $c$ denote the density and color predicted by $f_\phi$ at that point.
Once fully trained, $f_\phi$ can be used to render novel views of the 3D scene.
NeRF introduced \emph{positional encodings} of the input 3D coordinates, $\mathbf{x}$, to a higher dimensional space to capture high-frequency variations in color and geometry.
InstantNGP~\cite{mueller2022instant} further extended this idea to \textit{instance-specific} multi-resolution hash encodings (MRHE) to encode $\mathbf{x}$ dynamically based on scene properties.
These MRHEs, $h$, are learned along with the MLP parameters, $\phi$ for a given NeRF function, $f$ and show improved quality and reduced training/inference time.

\textbf{Image Denoising} 
is the process of reducing the noise and improving the perceptual quality of images while preserving important structural details.
Recent advancements in deep learning-based image restoration and denoising techniques~\cite{liang2021swinir, drcnet, mou2022dgun} have demonstrated remarkable success in removing noise and enhancing the perceptual quality of noisy images that may have suffered degradation.
Such networks are trained on large datasets of paired noisy and clean images to learn a mapping between the degraded input and the corresponding high-quality output by minimizing the difference between the restored and the ground truth clean image. In our case, we use denoising to improve the quality of our NeRF renderings by reducing artifacts and improving the texture and structure at the image level. 


\subsection{Method}

\begin{figure}[t]
    \centering 
    \includegraphics[width=0.9\linewidth]{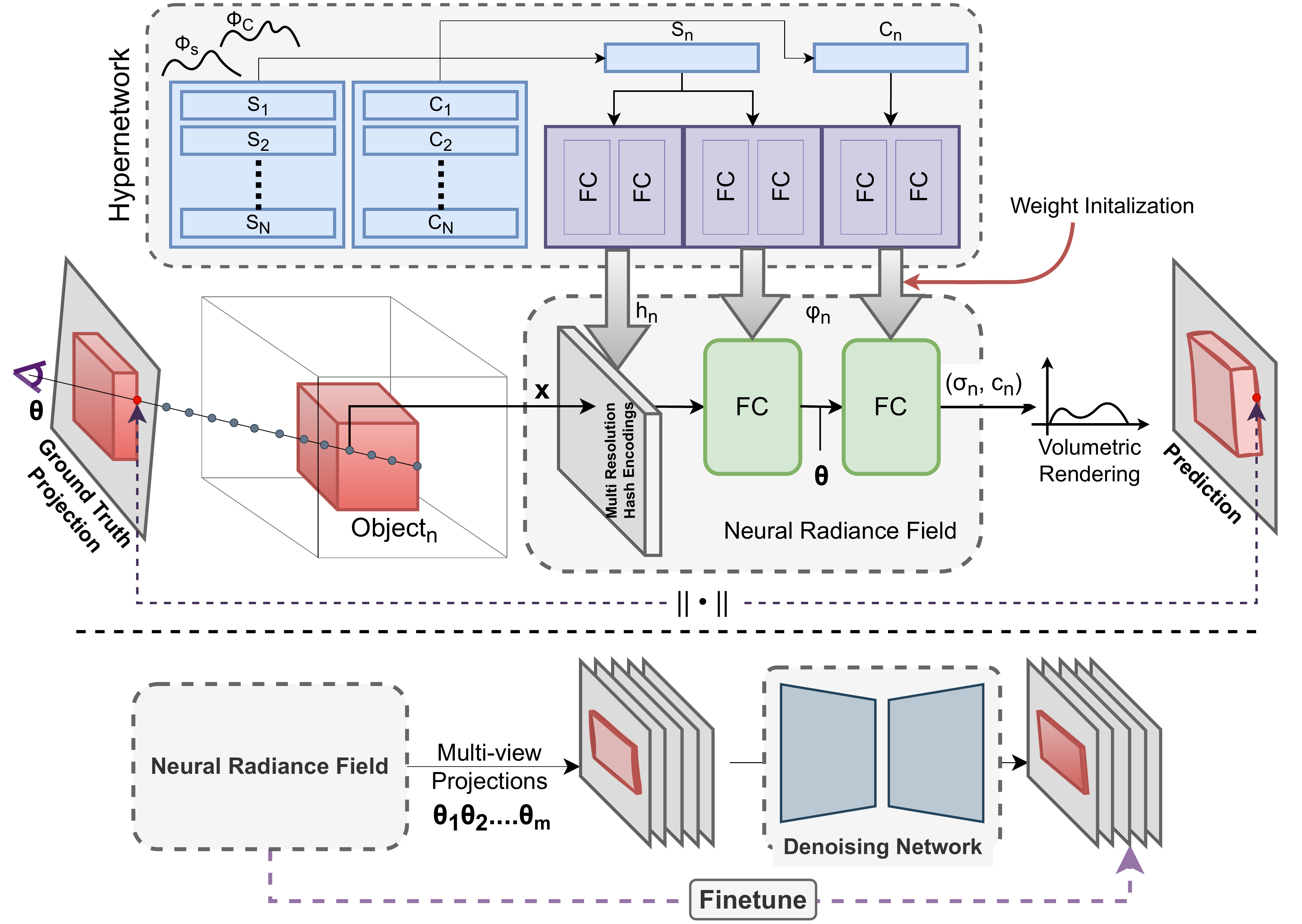}
    \caption{\textbf{Architecture Diagram: }\acro is trained and inferred in two steps.  
    In the first step \textbf{(top)}, our hypernetwork, $M$, is trained to predict the parameters of a NeRF model, $f_n$ corresponding to object instance $n$. At this stage, the NeRF model acts as a set of differentiable layers to compute the volumetric rendering loss, using which $M$ is trained on a set of $N$ objects, thereby learning a prior $\Phi$ = $\{\Phi_S, \Phi_C\}$ over the shape and color codes given by $S$ and $C$, respectively. 
    In the second step \textbf{(bottom)}, the quality of the predicted multiview consistent NeRF, $f_n$, is improved using a denoising network trained directly in the image space. To do this, $f_n$ is rendered from multiple known poses to a set of images that are improved to photorealistic quality. 
    $f_n$ is then finetuned on these improved images. 
    Importantly, since $f_n$ is only finetuned and not optimized from scratch, and thus $f_n$ retains the multiview consistency whilst improving in terms of texture and shape quality.
    } 
    \label{fig:architecture}
\end{figure}

Given a set of NeRFs denoted by $\{f_{(\phi_n, h_n)}\}_{n=1}^{N}$, where $N$ denotes the number of object instances in a given object category, we want to learn a prior $\Phi = \{\Phi_S, \Phi_C\}$, where $\Phi_S$ and $\Phi_C$ are the shape and color priors, respectively. Each NeRF, $f_{(\cdot)_n}$, is parameterized by the neural network weights, $\phi_n$, and learnable MRHEs, $h_i$ as proposed in \cite{mueller2022instant}. $f_{(\cdot)_n}$ takes a 3D position, $\mathbf{x}$, and viewing direction, $\theta$, as input and predicts the density conditioned on $\mathbf{x}$ denoted by $\sigma_n^{\{\mathbf{x}\}}$, and color conditioned on $\mathbf{x}$ and $\theta$ denoted by $c_n^{\{\mathbf{x}, \theta\}}$. This is given as,  
\begin{equation}
    f_{(\phi_n, h_n)}(\mathbf{x}, \theta) = \{\sigma_n^{\{\mathbf{x}\}}, c_n^{\{\mathbf{x}, \theta\}}\}.
\end{equation}
Our proposed method for learning NeRF priors involves two steps.
First, we train a hypernetwork, $M$, to learn a prior over a set of multiview consistent NeRFs of high-quality shape and texture.
Second, we employ an image-based denoising network that takes as input an already multiview consistent set of images, rendered from the predicted NeRF, and improves the shape and texture of NeRF to higher quality by finetuning on a set of denoised images.
Our architecture is outlined in \Cref{fig:architecture} and we explain each step in detail below.

\textbf{Step 1: Hypernetwork for Learning NeRF Prior.}
We want to design our hypernetwork, $M$, with trainable parameters, $\Omega$ that can predict NeRF parameters $\{\phi_n, h_n\}$ given a conditioning code $z_n = \{S_n, C_n\}$, where $S_n$ and $C_n$ are the shape and color codes, respectively, for an object instance $n$ belonging to a specific category.
Here, $S_n$ and $C_n$ belong to codebooks, $S$ and $C$ that are trained along with $\Omega$ in an auto-decoding fashion.

As shown in \Cref{fig:architecture}~(top), ideally we want $M$ to learn a prior $\{\Phi_C, \Phi_S\}$ over $S$ and $C$ such that given a random set of codes, $\{\mathcal{Y}_S \sim \Phi_S, \mathcal{Y}_C \sim \Phi_C\}$, $M$ should be able to generate a valid NeRF with consistent shape and texture for the given category of objects.  
To achieve this, we train $M$ by assuming the same constraints as are needed to train a NeRF - a set of multiview consistent images $\mathbf{I} = \{\{I_{\theta\in\Theta}\}_n\}_{n=1}^{N}$ for a set of poses, $\Theta$. In each training step, we start with a random object instance, $n$, and use the corresponding codes $S_n$ and $C_n$ from the codebooks as an input for $M$.
Our key insight is that estimating \textbf{both} the MRHEs and MLP weights results in a higher quality than other alternatives.
$M$ then predicts the NeRF parameters $\{\phi_n, h_n\}$, which is then used to minimize the following objective:
\begin{equation}
    \mathcal{L}(\Omega, S_n, C_n) = \sum_{\mathbf{r} \in R}||\mathbf{V}'(\mathbf{r}, \{\sigma_n^{\{x_i^\mathbf{r}\}}, c_{n}^{\{x_i^\mathbf{r}, \theta\}}\}_{i=1}^{L}) - \mathbf{V}_n(\mathbf{r})|| 
    \label{eqn:mainloss}
\end{equation}
\begin{equation}
    \{\sigma_n^{\{x_i^\mathbf{r}\}}, c_{n}^{\{x_i^\mathbf{r}, \theta\}}\} = f_{(\phi_n, h_n)}(x_i^\mathbf{r}, \theta) \quad \mathrm{and} \quad \{\phi_n, h_n\} = M_\Omega(S_n, C_n)
\end{equation}
where $\mathbf{V}'$ denotes the volumetric rendering function as given in \cite{nerf} eqn.~$3$ and $5$, $\mathbf{r}$ is a ray projected along the camera pose $\theta$, $x_i^\mathbf{r} \in \mathbf{x}$ and $L$ denote the number of points sampled along $\mathbf{r}$, and $\mathbf{V}_n$ denote the ground truth value for projection of the $n$\textsuperscript{th} object along $\mathbf{r}$.

Note that, in this step, the only trainable parameters are the meta-network  weights, $\Omega$, and the codebooks $S$ and $C$.
In this setting, the NeRF functions $f_{(\cdot)_n}$ only act as differentiable layers that allow backpropogation through to $M$ enabling it to train with multiview consistency loss attained by the volumetric rendering loss as described in \cite{nerf}. 
We use an instantiation of InstantNGP~\cite{mueller2022instant} as our function $f_{(\cdot)_n}$ consisting of MRHE and a small MLP. 

A general limitation of hypernetworks arises from the fact that the intended output space (i.e. the space of valid MLP weight matrices) is a subset of the actual output space, which is unristricted and can be any 2D matrix. Thus, a hypernetwork trained on loss functions in the weight space can result in unstable training, and might require a lot of training examples to converge. To overcome this issue, we train our hypernetwork end-to-end directly on images, so that it learns the implicit NeRF space along with the category specific prior on it, which simplifies the setting for the hypernetwork and allows for more stable training. As a causal effect of this, HyP-NeRF, when trained on less number of examples, essentially acts as a compressing model.

\textbf{Step 2: Denoise and Finetune.}
\label{sec:denoise}
In the first step, $M$ is trained to produce a consistent NeRF with high-fidelity texture and shape. However, we observed that there is room to improve the generated NeRFs to better capture fine details like uneven textures and edge definition. 
To tackle this challenge, we augment $M$ using a denoising process that takes $f_{(\cdot)_n}$ and further finetunes it to achieve $f_{(\cdot)_n}^H$. 

As shown in \Cref{fig:architecture}~(bottom), we render novel views from the multiview consistent NeRF into $m$ different predefined poses given by $\{\theta_1, \theta_2...\theta_m\}$ to produce a set of multiview consistent images $\{\hat{I}_i\}_{i=1}^{m}$.
We then use a pre-trained image-level denoising autoencoder that takes $\{\hat{I}_i\}_{i=1}^{m}$ as input and produces images of improved quality given as $\{\hat{I}^H_i\}_{i=1}^m$. 
These improved images are then used to \underline{finetune} $f_{(\cdot)_n}$ to achieve $f_{(\cdot)_n}^H$.
Note that, we do not train the NeRFs from scratch on $\{\hat{I}^H\}$ and only finetune the NeRFs, which ensures fast optimization and simplifies the task of the denoising module that only needs to improve the quality and does not necessarily need to maintain the multiview consistency.
While our denoising is image-level, we still obtain multiview consistent NeRFs since we finetune on the NeRF itself (as we also demonstrate through experiments in the Appendix).

For our denoising autoencoder, we use VQVAE2~\cite{vqvae2} as the backbone. To train this network, we simply use images projected from the NeRF, predicted by the hypernetwork (lower quality relative to the ground truth) as the input to the VQVAE2 model. We then train VQVAE2 to decode the ground truth by minimizing the L2 loss objective between VQVAE2's output and the ground truth.

\subsection{\acro Inference and Applications}
\label{sec:query_net}
Training over many NeRF instances, $M$ learns a prior $\Phi$ that can be used to generate novel consistent NeRFs. However, $\Phi$ is not a known distribution like Gaussian distributions that can be naively queried by sampling a random point from the underlying distribution. We tackle this in two ways:

\label{sec:tto}
\textbf{Test Time Optimization}. In this method, given a single-view or multi-view posed image(s), we aim to estimate shape and color codes $\{S_o, C_o\}$ of the NeRF that renders the view(s). To achieve this, we freeze $M$'s parameters and optimize the $\{S_o, C_o\}$ using the objective given in \Cref{eqn:mainloss}. 

\textbf{Query Network}. We create a query network, $\Delta$, that maps a point from a known distribution to $\Phi$. As CLIP's~\cite{clip} pretrained semantic space, say $\mathbf{C}$, is both text and image aware, we chose $\mathbf{C}$, as our known distribution and learn a mapping function $\Delta(z \sim \mathbf{C}) \to \Phi$. Here, $\Delta$ is an MLP that takes $z$ as input and produces $\mathcal{Y}_z \in \Phi$ as output. To train $\Delta$, we randomly sample one pose from the ground truth multiview images $I_\theta^n \in \{I_{\theta \in \Theta}\}_n$ and compute the semantic embedding $z_\theta^n=\textrm{CLIP}(I_\theta^n)$ and map it to $\{\bar{S}_n, \bar{C}_n\} \in \Phi$ given as $\{\bar{S}_n, \bar{C}_n\} = \Delta(z_\theta^n)$. We then train our query network by minimizing the following objective: 
\vspace{-0.1in}
\begin{equation}
    \mathcal{L}_\Delta = \sum_\theta||\{\bar{S}_n, \bar{C}_n\}, \{S_n, C_n\}||. 
\end{equation}
\vspace{-0.22in}

At the time of inference, given a text or image modality such as a text prompt, single-view unposed (in-the-wild) image, or segmented image, we compute the semantic embedding using CLIP encoder and map it to $\Phi$ using $\Delta$, 
from which we obtain the shape and color codes as input for the \acro.  

Note that for N query points in a scene, the forward pass through the hypernetwork (computationally expensive) happens only once per scene. Only the NeRF predicted by the hypernetwork (computationally less expensive) is run for each query point. 

\section{Experiments}

\begin{table}[t]
    \centering
    \adjustbox{max width=\linewidth}{
    \begin{tabular}
    {r|l|cccc|cccc}
    \toprule
    & & \multicolumn{4}{c|}{Chairs} & \multicolumn{4}{c}{Sofa} \\
    & & PSNR$\uparrow$ & SSIM$\uparrow$ & LPIPS$\downarrow$& FID$\downarrow$& PSNR$\uparrow$ & SSIM$\uparrow$ & LPIPS$\downarrow$& FID$\downarrow$ \\
    \midrule
    \multirow{3}{*}{ABO-512} & PixelNeRF~\cite{pixelnerf} & 18.30 & 0.83 & 0.31 & 292.32 & 17.51 & 0.84 & 0.28 & 323.89 \\
    & CodeNeRF~\cite{jang2021codenerf} & 19.86 & 0.87 & 0.298 & - & 19.56 & 0.87 & 0.290 & - \\
    & \acro (Ours) & \textbf{24.23} & \textbf{0.91} & \textbf{0.16} & \textbf{68.11} & \textbf{23.96} & \textbf{0.90} & 0.18 & \textbf{120.80} \\ 
    & \hspace{1cm} w/o Denoise & 23.05 & 0.90 & \textbf{0.16 }& 102.45 & 23.54 & \textbf{0.90} & \textbf{0.174} & 121.69 \\
    \bottomrule
    \end{tabular}}
    \vspace{0.1cm}
    \caption{\textbf{Generalization}. Comparison of single-posed-view NeRF generation. 
    Metrics are computed on renderings of resolution $512\times512$. \acro significantly outperforms PixelNeRF and CodeNeRF on all the metrics in both the datasets. }
    \label{tab:gen_exps}
    \vspace{-0.6cm}
\end{table}

We provide evaluations of the prior learned by \acro specifically focusing on the quality of the generated NeRFs.
We consider three dimensions: (1) \textbf{Generalization}~(\Cref{sec:generalization}): we validate whether \acro can generate novel NeRFs not seen during training by conditioning on only a single-posed-view of \underline{novel} NeRF instances.
(2) \textbf{Compression} (\Cref{sec:compression}): since \acro is trained in an auto-decoding fashion on specific NeRF instances (see \Cref{eqn:mainloss}), we can evaluate the quality of the NeRFs compressed in this process.
(3) \textbf{Retrieval} (\Cref{sec:retrieval}): as shown in \Cref{fig:applications}, \acro's prior enables various downstream applications. We show how to combine our prior with CLIP~\cite{clip} to retrieve novel NeRFs.

\begin{figure}[!t]
    \centering 
    \includegraphics[width=\linewidth]{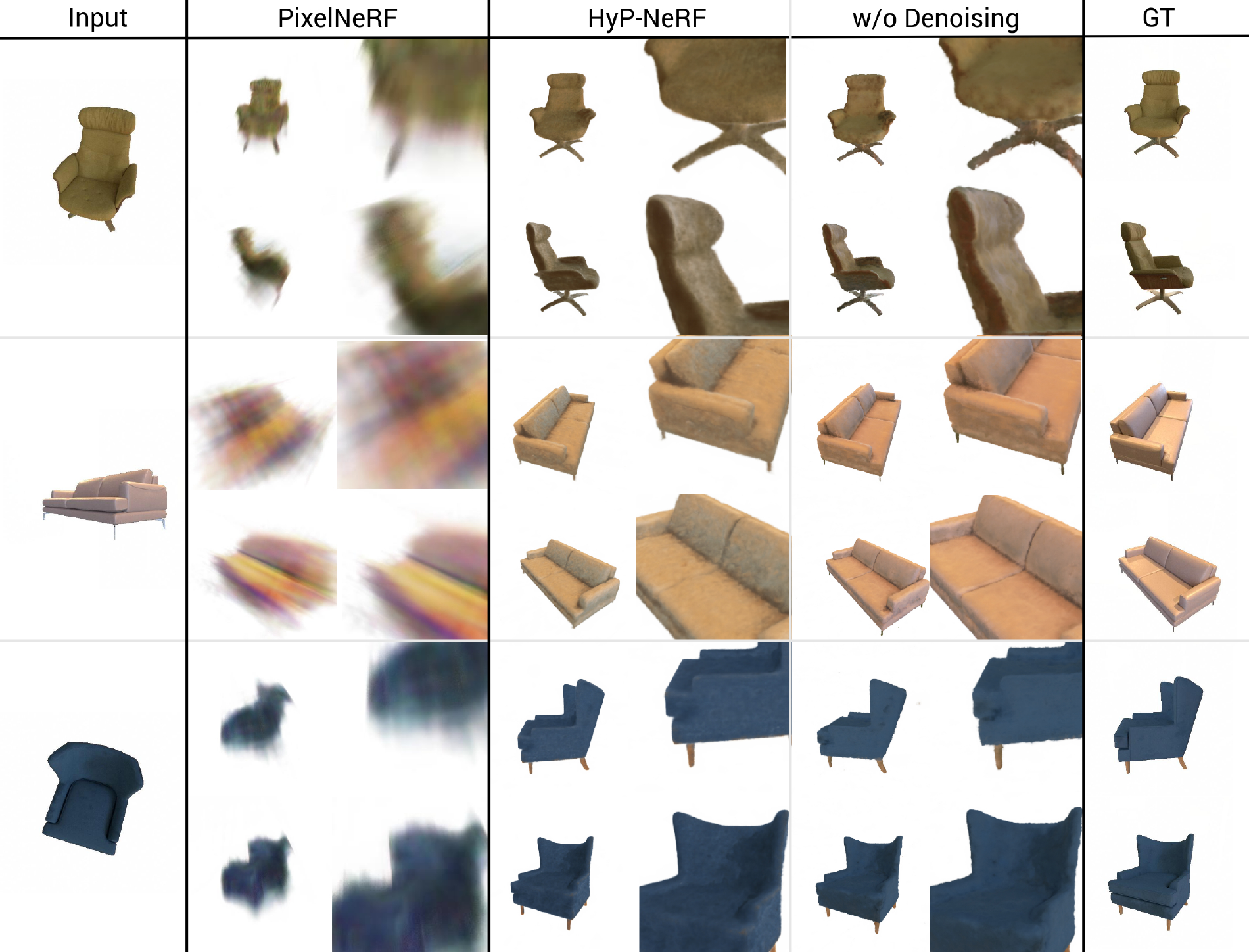}
    \vspace{-1em}
    \caption{\textbf{Qualitative Comparison of Generalization on ABO.} The NeRFs are rendered at a resolution of $512\times512$. \acro is able to preserve fine details such as the legs, creases, and texture even for novel instances. PixelNeRF fails to preserve details and to model the structure.}
    \label{fig:visual-abo}
\end{figure}

\begin{table}[t]
    \centering
    \adjustbox{width=\linewidth}{
    \begin{tabular}{r|cccc|cccc|cccc}
    \toprule
    & \multicolumn{4}{c|}{ABO Chairs} & \multicolumn{4}{c|}{ABO Table} & \multicolumn{4}{c}{ABO Sofas} \\
    & PSNR$\uparrow$ & SSIM$\uparrow$ & LPIPS$\downarrow$ & CD $\downarrow$ &
    PSNR$\uparrow$ & SSIM$\uparrow$ & LPIPS$\downarrow$ & CD $\downarrow$ &
    PSNR$\uparrow$ & SSIM$\uparrow$ & LPIPS$\downarrow$ & CD $\downarrow$ \\
    \toprule
         \cite{mueller2022instant} &  35.43 & 0.96 &0.07 & $-$ & 34.07 & 0.95 &0.07 & $-$  &33.87 &0.95 &0.08 & $-$ \\
         Ours & 31.37 & 0.94 & 0.1 & 0.0082 &29.52 &0.93&0.11& 0.0033 & 30.32& 0.94 & 0.11 & 0.0118 \\
    \bottomrule
    \end{tabular}}
    \vspace{0.1cm}
    \caption{\textbf{Compression}. We randomly sample $250$ datapoints from our training dataset and compare the NeRFs learned using InstantNGP~\cite{mueller2022instant} on the individual instances against \acro that learns the entire dataset. \underline{Note}, we do not employ the denoising module (see \Cref{sec:denoise}) for this evaluation.} 
    \label{tab:compression}
    \vspace{-0.6cm}
\end{table}

\textbf{Datasets and Comparisons}. We primarily compare against two baselines, PixelNeRF~\cite{pixelnerf} and InstantNGP~\cite{mueller2022instant} on the Amazon-Berkeley Objects (ABO)~\cite{abo} dataset.
ABO contains diverse and detailed objects rendered at a resolution of $512\times512$ which is perfect to showcase the quality of the NeRF generated by \acro.
Rather than use a computationally expensive model like VisionNeRF (on the SRN~\cite{3dhypernet2} dataset) on a resolution of $128\times128$, we show our results on $512\times512$ and compare with PixelNeRF.
Additionally, we compare with the other baselines on SRN at $128\times128$ resolution qualitatively in the main paper (\Cref{fig:srn_quant}) and quantitatively in the Appendix.
For compression, we directly compare with InstantNGP~\cite{mueller2022instant}, that proposed MRHE, trained to fit on individual objects instance-by-instance.

\begin{wraptable}[9]{R}{5cm}
    \vspace{-.9em}
    \centering
    \adjustbox{width=\linewidth}{
    \begin{tabular}{cc|cc}
    \toprule
    \multicolumn{2}{c|}{ABO Chairs} & \multicolumn{2}{c}{ABO Sofa} \\
    Top 1 & Top 3 & Top 1 & Top 3\\
    \toprule    
    98.72\% & 99.81\% & 91.6\% & 95.27\% \\
    \bottomrule
    \end{tabular}}
    \vspace{-0.2cm}
    \caption{\small \textbf{Retrieval}. We design a simple query network (see \Cref{sec:query_net}) to retrieve NeRF instances from \acro's prior seen at the time of training and achieve almost 100\% accuracy.}
    \label{tab:retrieval}
\end{wraptable}

\textbf{Architectural Details}.
We use InstantNGP as $f_{(\cdot)_n}$, with $16$ levels, hashtable size of $2^{11}$, feature dimension of $2$, and linear interpolation for computing the MRHE; the MLP has a total of 5, 64-dimensional, layers. We observed that a hashtable size $2^{11}$ produces NeRF of high-quality at par with the a size of $2^{14}$. Hence, we use $2^{11}$ to speed up our training. Our hypernetwork, $M$, consists of 6 MLPs, 1 for predicting the MRHE, and the rest predicts the parameters $\phi$ for each of the MLP layers of $f$. Each of the MLPs are made of 3, 512-dimensional, layers. We perform all of our experiments on NVIDIA RTX 2080Tis. 

\textbf{Metrics}.
To evaluate NeRF quality, we render them at 91 distinct views and compute metrics on the rendered images.
Following PixelNeRF, we use PSNR($\uparrow$), SSIM($\uparrow$), and LPIPS($\downarrow$)~\cite{lpips}.
Additionally, we compute Fréchet Inception Distance (FID)($\downarrow$)~\cite{heusel2017gans} to further test the visual quality.
Although these metrics measure the quality of novel-view synthesis, they do not necessarily evaluate the geometry captured by the NeRFs.
Therefore, we compute Chamfer's Distance (CD) whenever necessary by extracting a mesh from NeRF densities~\cite{lorensen1987marching}. 
Please see the Appendix for additional details. 

\subsection{Generalization}
\label{sec:generalization}
One way to evaluate if \acro can render novel NeRF instances of high quality is through unconditional sampling.
However, our learned prior $\Phi$ is a non-standard prior (like a Gaussian distribution) and thus random sampling needs carefully designed mapping between such a known prior and $\Phi$.
Therefore, we instead rely on a conditional task of single-view novel NeRF generation: given a single arbitrarily-chosen view of a novel object, we generate the corresponding NeRF, $f_{(\cdot)_o}$ through test-time optimization (see \Cref{sec:tto}). We compare quantitatively with PixelNeRF on ABO at a high resolution of $512\times512$ and qualitatively with the rest of the baselines on SRN at $128\times128$. 

\begin{figure}
    \centering 
    \includegraphics[width=\linewidth]{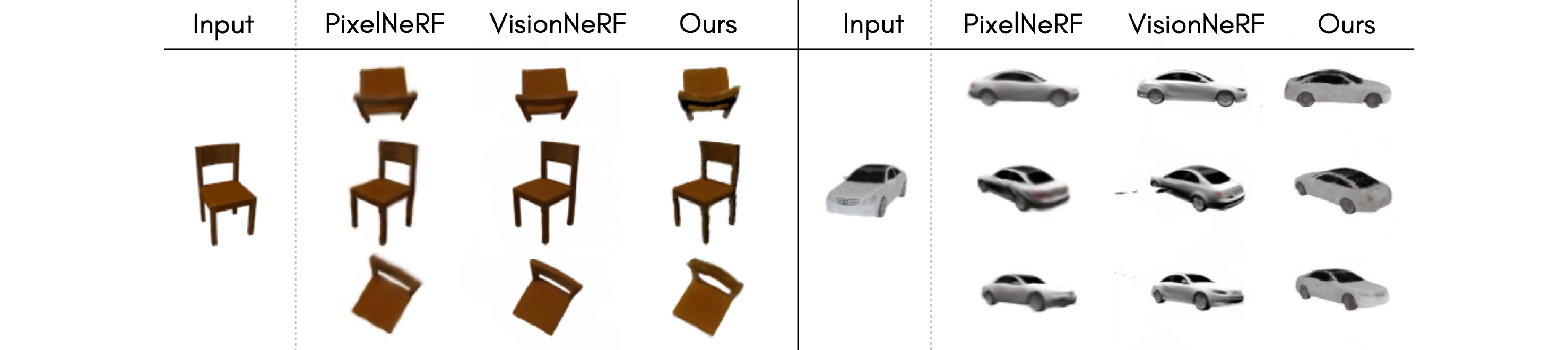}
    \vspace{-1.3em}
    \caption{\textbf{Qualitative Comparison of Generalization on SRN} on the task of single-view inversion (posed in our case) and compare the quality of the views rendered at $128\times128$. 
    \acro renders NeRFs of similar quality to the PixelNeRF and VisionNeRF baselines. }
    \label{fig:srn_quant}
    \vspace{-0.6cm}
\end{figure}

As shown in \Cref{tab:gen_exps}, we significantly outperform PixelNeRF on all of the metrics. Further, the qualitative results in \Cref{fig:visual-abo} clearly shows the difference between the rendering quality of \acro against PixelNeRF.
Specifically, PixelNeRF fails to learn details, especially for the Sofa category.
On the other hand, \acro preserves intricate details like the texture, legs, and folds in the objects even at a high resolution. 
Further, we show our results on the widely used SRN dataset at the resolution of $128\times128$ in \Cref{fig:srn_quant}.
Here, our quality is comparable with the baselines.

\begin{wrapfigure}[23]{R}{7.5cm}
    \vspace{-2.3em}
    \centering 
    \includegraphics[width=\linewidth]{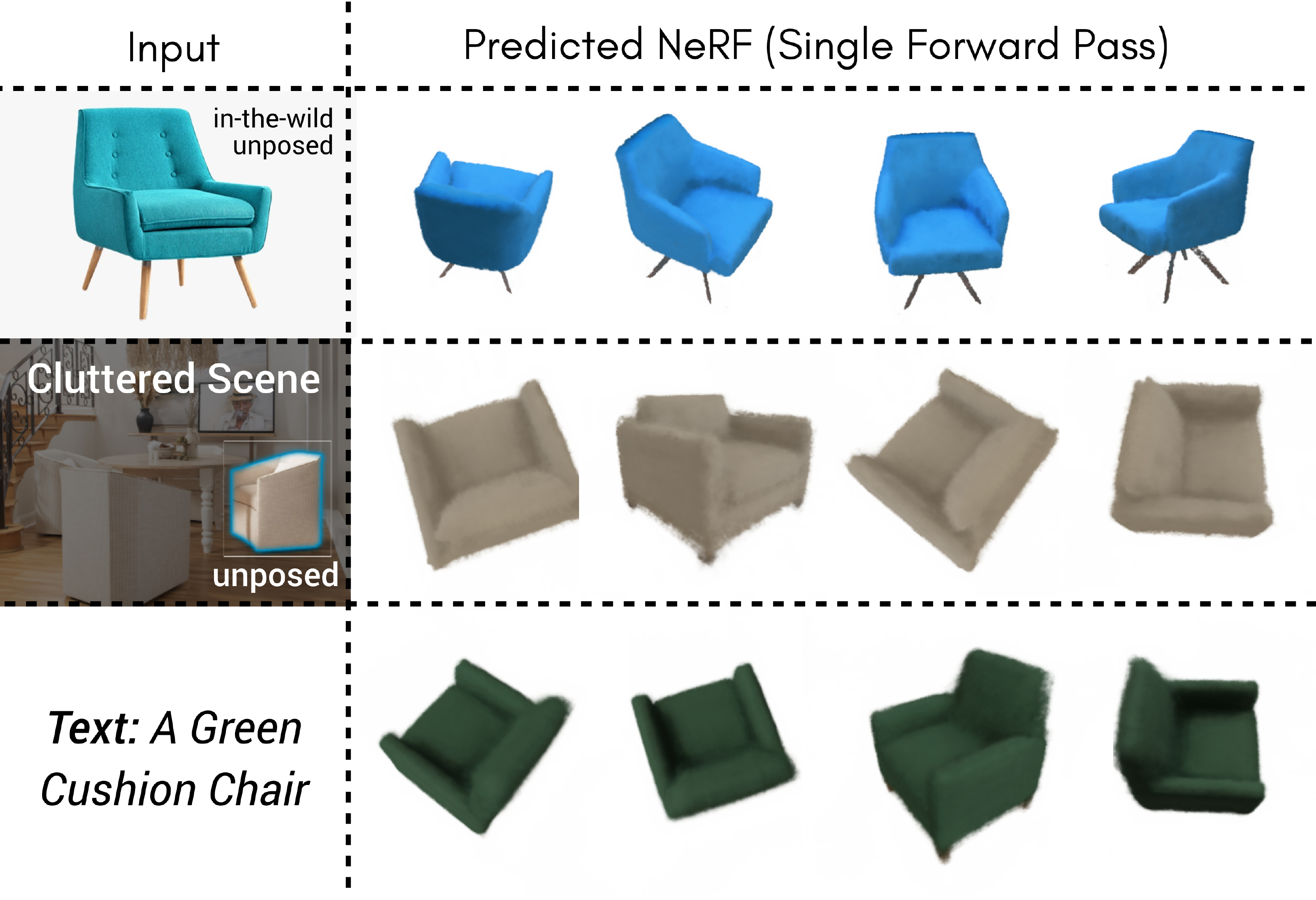}
    \vspace{-1.8em}
    \caption{\textbf{Qualitative Comparison of Querying (\Cref{sec:query_net}) on \acro's prior}. In the top, we use an in-the-wild single-view unposed image to retrieve the closest NeRF \acro has seen during training. In middle, we take a cluttered scene, and mask out the object of interest using Segment Anything~\cite{kirillov2023segany} and in the bottom we use a text prompt as an input to our query network, $\Delta$. We then obtain the latent codes $\{S, C\}$ from $\Delta$, which are used as an input for \acro.
    }
    \label{fig:retrieval}
\end{wrapfigure}

\subsection{Compression}
\label{sec:compression}
Unlike InstantNGP, which is trained on a single 3D instance, \acro is trained on many NeRF instances which effectively results in the compression of these NeRFs into the latent space (or the codebook).
We evaluate this compression capability by computing NeRF quality degradation compared to single-instance-only method, InstantNGP.

We randomly sample 250 instances from the training set and train InstantNGP separately on each of them.
These samples are a subset of the training data used in \acro's codebook.
We show degradation metrics in \Cref{tab:compression}.
Note that we \textbf{do not perform denoising} on the generated NeRFs as we want to only evaluate the compression component of \acro in this section.
As can be seen in \Cref{tab:compression}, there is a significant degradation in terms of PSNR (an average of $11\%$), but the overall geometry is preserved almost as well as InstantNGP.
However, InstantNGP is trained on a single instance, whereas we train on 1000s of NeRF instances (1038, 783, and 517 instances for ABO Chairs, Sofa, and Tables, respectively).
This results in a 60$\times$ compression gain: 
for ABO Chairs, with 1038 training instances, \acro needs 163MB to store the model, whereas a single instance of InstantNGP needs on average 8.9MB.
Note that we use the same network architecture~\cite{githubGitHubAshawkeytorchngp} for \acro and InstantNGP making this a fair comparison.
Moreover, the storage complexity for InstantNGP-based NeRFs is linear with respect to the number of instances, whereas our degradation in visual quality is sublinear.

\vspace{-8px}
\subsection{Retrieval} 
\label{sec:retrieval}
A generalizable learned prior has the ability to generate NeRFs based on different input modalities like text, images, segmented and occluded images, random noise, and multi-view images. 
We now demonstrate additional querying and retrieval capabilities as described in \Cref{sec:query_net}.

This experiment's goal is to retrieve specific NeRF instances that \acro has encountered during training from a single-view unposed image of that instance. 
\Cref{sec:retrieval} presents the number of times we could correctly retrieve from an arbitrary view of seen NeRF instances.
We achieve almost 100\% accuracy for Chair and Sofa datasets.
However, we take this a step further and try to retrieve the closest training instance code corresponding to \textbf{unseen views} of seen instances taken from in-the-wild internet images. 
\Cref{fig:retrieval} (top) shows examples from this experiment in which we are able to retrieve a NeRF closely matching the input query. 
This demonstrates the ease of designing a simple mapping network that can effectively interact with \acro's prior. 

Along with retrieving a seen instance, we use the query network to generate novel NeRFs of \textbf{unseen instances} as shown in \Cref{fig:retrieval} (middle and bottom).
In the middle row, we take an image of a cluttered scene, segment it with SAM~\cite{kirillov2023segany}, and pass this as input to the query network, from which we obtain a set of latent codes given as input to \acro (see \Cref{fig:applications}).
Finally, in the bottom row, we show text-to-NeRF capabilities enabled by \acro.

\subsection{Ablation} 
\label{sec:ablation}

\begin{wraptable}[]{R}{6cm}
    \vspace{-0.8em}
    \centering
    \adjustbox{width=\linewidth}{
    \begin{tabular}{r|cccc}
    \toprule
    & \multicolumn{4}{c}{Chairs} \\
    & PSNR$\uparrow$ & SSIM$\uparrow$ & LPIPS$\downarrow$ & CD$\downarrow$ \\
    \toprule
    \acro & \textbf{29.23} & \textbf{0.94} & \textbf{0.10} & \textbf{0.0075} \\  
    w/o MRHE & 26.42 & 0.92 & 0.16 & 0.0100\\  
    \bottomrule
    \end{tabular}}
    \vspace{0.05cm}
    \caption{\small \textbf{Ablation of removing MRHE} on ABO dataset. Due to the significant rendering time of \acro w/o MRHE, we sample 70 object instances from the training dataset to compute the metrics at $512\times512$ resolution.}
    \label{tab:ablation}
\end{wraptable}

Two key designs of \acro include incorporating the MRHE and the denoising network. We present the affect of removing these two components in \Cref{tab:ablation} and \Cref{fig:teaser} for MRHE and \Cref{tab:gen_exps}, \Cref{fig:applications}, and \Cref{fig:visual-abo} for denoising. In the first ablation, we change the design of our neural network by using a hypernetwork to predict the parameters of a standard nerf with positional encodings~\cite{nerf}. Since we remove the MRHE, we also increase the number of layers in the MLP to match the layers mentioned in \cite{nerf}.
Since there is a significant increase in the view rendering time, we randomly sample 70 training examples for evaluation.
As seen in \Cref{tab:ablation}, the quality of the rendered views lags significantly in all the metrics including the CD (measured against NeRFs rendered on InstantNGP individually).
This is showcased visually in \Cref{fig:teaser} and the Appendix.
Similarly, we find significant differences between the quality of the NeRFs before and after denoising (\Cref{tab:gen_exps}, \Cref{fig:applications}, and \Cref{fig:visual-abo}), particularly in the Chair category with more diverse shapes. 

\subsection{Color and Shape Disentanglement}
\label{sec:color_shape_disentanglement}
\begin{figure}[t]
    \centering
    \includegraphics[width=\linewidth]{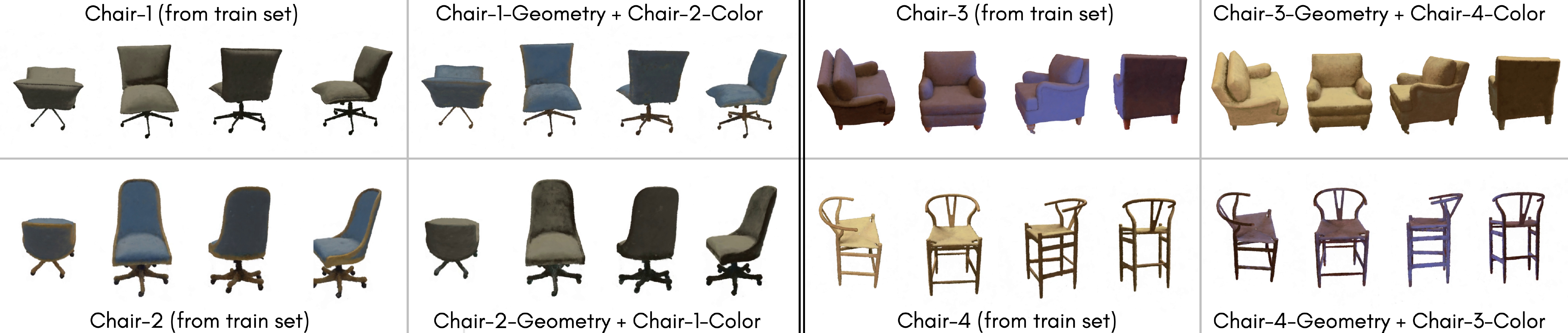}
    \vspace{-12px}
    \caption{Qualitative results on color and geometry disentanglement: We take two instances from the training set and switch the geometry and color codes to generate novel instances. As can be seen, the geometry and the color are transferred while preserving fine shape details. Even the fine details, like stripes and color-contrast between the chair seats and edges from Chair-2, are accurately transferred to Chair-1 (Chair-2-Geometry + Chair-1-Color). \textbf{Zoom in for an improved experience.}}
    \label{fig:color_geometry_disentanglementl}
\end{figure}

\begin{figure}[]
    \centering
    \includegraphics[width=\linewidth]{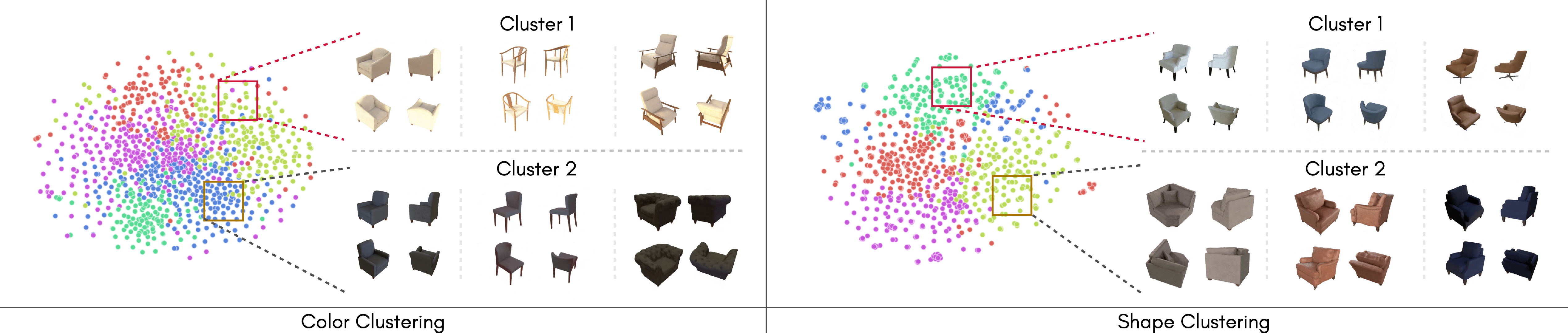}
    \vspace{-14px}
    \caption{Latent visualization through TSNE plots on shape and color codes (from the codebooks $S$ and $C$). As can be seen, the underlying space forms meaningful clusters as shown through examples randomly sampled from two different clusters. }
    \label{fig:tsne_plots}
\end{figure}

We start with two object instances from the train set - \textit{A} and \textit{B} and denote their corresponding shape and color codes as \textit{$A_s$}, \textit{$B_s$} and \textit{$A_c$}, \textit{$B_c$}. We switch the color and shape code and generate two novel NeRFs given by \textit{{$A_s$, $B_c$}} and \textit{{$B_s$, $A_c$}}. In \Cref{fig:color_geometry_disentanglementl}, we can clearly see the disentanglement: geometry is perfectly preserved, and the color is transferred faithfully across the NeRFs. In \Cref{fig:tsne_plots}, we show clusters of color and shape codes using TSNE plots and visualize instances from the clusters. 

\section{Conclusion, Limitation, and Future Work}
We propose \acro, a learned prior for Neural Radiance Fields (NeRFs).
\acro uses a hypernetwork to predict instance-specific multi-resolution hash encodings (MRHEs) that significantly improve the visual quality of the predicted NeRFs.
To further improve the visual quality, we propose a denoising and finetuning technique that result in an improved NeRF that preserves its original multiview and geometric consistency.  
Experimental results demonstrate \acro's capability to generalize to unseen samples and its effectiveness in compression. 
With its ability to overcome limitations of existing approaches, such as rendering at high resolution and multiview consistency, \acro holds promise for various applications as we demonstrate for single- and multi-view NeRF reconstruction and text-to-NeRF. 

\textbf{Limitation and Future Work}. 
One limitation of our work is the need for the pose to be known during test-time optimization (\Cref{sec:tto}). Although we propose the query network to predict novel NeRFs conditioned on an unposed single view, the result may not exactly match the given view because of the loss of detail in the CLIP embedding. 
Future work should design a mapping network that can preserve fine details. 
An iterative pose refinement approach that predicts the pose along with the shape and color codes could also be adopted.
A second limitation of our work is the non-standard prior $\Phi$ that was learned by \acro which makes unconditional generation challenging. 
GAN-based generative approaches solve this problem by randomly sampling from a standard distribution (like Gaussian distribution) and adversarially training the network.
However, those methods often focus more on image quality than 3D structure.
Future work could address this by incorporating latent diffusion models that can map a standard prior to \acro's prior.

\bibliography{submission}

\begin{thebibliography}{10}

\bibitem{githubGitHubAshawkeytorchngp}
{G}it{H}ub - ashawkey/torch-ngp: {A} pytorch {C}{U}{D}{A} extension
  implementation of instant-ngp (sdf and nerf), with a {G}{U}{I}. ---
  github.com.
\newblock \url{https://github.com/ashawkey/torch-ngp}.
\newblock [Accessed 17-May-2023].

\bibitem{mipnerf}
Jonathan~T. Barron, Ben Mildenhall, Matthew Tancik, Peter Hedman, Ricardo
  Martin-Brualla, and Pratul~P. Srinivasan.
\newblock Mip-nerf: A multiscale representation for anti-aliasing neural
  radiance fields.
\newblock {\em ICCV}, 2021.

\bibitem{barron2022mipnerf360}
Jonathan~T. Barron, Ben Mildenhall, Dor Verbin, Pratul~P. Srinivasan, and Peter
  Hedman.
\newblock Mip-nerf 360: Unbounded anti-aliased neural radiance fields.
\newblock {\em CVPR}, 2022.

\bibitem{pix2nerf}
Shengqu Cai, Anton Obukhov, Dengxin Dai, and Luc Van~Gool.
\newblock Pix2nerf: Unsupervised conditional p-gan for single image to neural
  radiance fields translation.
\newblock In {\em Proceedings of the IEEE/CVF Conference on Computer Vision and
  Pattern Recognition (CVPR)}, pages 3981--3990, June 2022.

\bibitem{fwd}
Ang Cao, Chris Rockwell, and Justin Johnson.
\newblock Fwd: Real-time novel view synthesis with forward warping and depth.
\newblock {\em CVPR}, 2022.

\bibitem{pigan}
Eric Chan, Marco Monteiro, Petr Kellnhofer, Jiajun Wu, and Gordon Wetzstein.
\newblock pi-gan: Periodic implicit generative adversarial networks for
  3d-aware image synthesis.
\newblock In {\em Proc. CVPR}, 2021.

\bibitem{eg3d}
Eric~R. Chan, Connor~Z. Lin, Matthew~A. Chan, Koki Nagano, Boxiao Pan,
  Shalini~De Mello, Orazio Gallo, Leonidas Guibas, Jonathan Tremblay, Sameh
  Khamis, Tero Karras, and Gordon Wetzstein.
\newblock Efficient geometry-aware {3D} generative adversarial networks.
\newblock In {\em arXiv}, 2021.

\bibitem{ssdnerf}
Hansheng Chen, Jiatao Gu, Anpei Chen, Wei Tian, Zhuowen Tu, Lingjie Liu, and
  Hao Su.
\newblock Single-stage diffusion nerf: A unified approach to 3d generation and
  reconstruction.
\newblock In {\em ICCV}, 2023.

\bibitem{garf}
Shin-Fang Chng, Sameera Ramasinghe, Jamie Sherrah, and Simon Lucey.
\newblock Garf: gaussian activated radiance fields for high fidelity
  reconstruction and pose estimation.
\newblock {\em arXiv e-prints}, pages arXiv--2204, 2022.

\bibitem{mou2022dgun}
Jian~Zhang Chong~Mou, Qian~Wang.
\newblock Deep generalized unfolding networks for image restoration.
\newblock In {\em IEEE Conference on Computer Vision and Pattern Recognition
  (CVPR)}, 2022.

\bibitem{abo}
Jasmine Collins, Shubham Goel, Kenan Deng, Achleshwar Luthra, Leon Xu, Erhan
  Gundogdu, Xi~Zhang, Tomas~F Yago~Vicente, Thomas Dideriksen, Himanshu Arora,
  Matthieu Guillaumin, and Jitendra Malik.
\newblock Abo: Dataset and benchmarks for real-world 3d object understanding.
\newblock {\em CVPR}, 2022.

\bibitem{hncompress2}
Shangqian Gao, Feihu Huang, and Heng Huang.
\newblock Model compression via hyper-structure network, 2021.

\bibitem{gu2023nerfdiff}
Jiatao Gu, Alex Trevithick, Kai-En Lin, Josh Susskind, Christian Theobalt,
  Lingjie Liu, and Ravi Ramamoorthi.
\newblock Nerfdiff: Single-image view synthesis with nerf-guided distillation
  from 3d-aware diffusion.
\newblock In {\em International Conference on Machine Learning}, 2023.

\bibitem{guo2022fast}
Pengsheng Guo, Miguel~Angel Bautista, Alex Colburn, Liang Yang, Daniel
  Ulbricht, Joshua~M Susskind, and Qi~Shan.
\newblock Fast and explicit neural view synthesis.
\newblock In {\em Proceedings of the IEEE/CVF Winter Conference on Applications
  of Computer Vision}, pages 3791--3800, 2022.

\bibitem{hypernetwork}
David Ha, Andrew Dai, and Quoc~V Le.
\newblock Hypernetworks.
\newblock {\em arXiv preprint arXiv:1609.09106}, 2016.

\bibitem{instructnerf2nerf}
Ayaan Haque, Matthew Tancik, Alexei Efros, Aleksander Holynski, and Angjoo
  Kanazawa.
\newblock Instruct-nerf2nerf: Editing 3d scenes with instructions.
\newblock 2023.

\bibitem{heusel2017gans}
Martin Heusel, Hubert Ramsauer, Thomas Unterthiner, Bernhard Nessler, and Sepp
  Hochreiter.
\newblock Gans trained by a two time-scale update rule converge to a local nash
  equilibrium.
\newblock {\em Advances in neural information processing systems}, 30, 2017.

\bibitem{jain2022zero}
Ajay Jain, Ben Mildenhall, Jonathan~T Barron, Pieter Abbeel, and Ben Poole.
\newblock Zero-shot text-guided object generation with dream fields.
\newblock In {\em Proceedings of the IEEE/CVF Conference on Computer Vision and
  Pattern Recognition}, pages 867--876, 2022.

\bibitem{jang2021codenerf}
Wonbong Jang and Lourdes Agapito.
\newblock Codenerf: Disentangled neural radiance fields for object categories.
\newblock In {\em Proceedings of the IEEE/CVF International Conference on
  Computer Vision}, pages 12949--12958, 2021.

\bibitem{stylegan2-ada}
Tero Karras, Miika Aittala, Janne Hellsten, Samuli Laine, Jaakko Lehtinen, and
  Timo Aila.
\newblock Training generative adversarial networks with limited data.
\newblock {\em ArXiv}, abs/2006.06676, 2020.

\bibitem{stylegan}
Tero Karras, Samuli Laine, and Timo Aila.
\newblock A style-based generator architecture for generative adversarial
  networks.
\newblock In {\em Proceedings of the IEEE/CVF conference on computer vision and
  pattern recognition}, pages 4401--4410, 2019.

\bibitem{kirillov2023segany}
Alexander Kirillov, Eric Mintun, Nikhila Ravi, Hanzi Mao, Chloe Rolland, Laura
  Gustafson, Tete Xiao, Spencer Whitehead, Alexander~C. Berg, Wan-Yen Lo, Piotr
  Doll{\'a}r, and Ross Girshick.
\newblock Segment anything.
\newblock {\em arXiv:2304.02643}, 2023.

\bibitem{hnfewshot2}
A.~Lamb, Evgeny~S. Saveliev, Yingzhen Li, Sebastian Tschiatschek, Camilla
  Longden, Simon Woodhead, Jos{\'e}~Miguel Hern{\'a}ndez-Lobato, Richard~E.
  Turner, Pashmina Cameron, and Cheng Zhang.
\newblock Contextual hypernetworks for novel feature adaptation.
\newblock {\em ArXiv}, abs/2104.05860, 2021.

\bibitem{drcnet}
Fei Li, Lingfeng Shen, Yang Mi, and Zhenbo Li.
\newblock Drcnet: Dynamic image restoration contrastive network.
\newblock In Shai Avidan, Gabriel Brostow, Moustapha Ciss{\'e}, Giovanni~Maria
  Farinella, and Tal Hassner, editors, {\em Computer Vision -- ECCV 2022},
  pages 514--532, Cham, 2022. Springer Nature Switzerland.

\bibitem{nerfpose}
Fu~Li, Hao Yu, Ivan Shugurov, Benjamin Busam, Shaowu Yang, and Slobodan Ilic.
\newblock Nerf-pose: A first-reconstruct-then-regress approach for
  weakly-supervised 6d object pose estimation.
\newblock {\em arXiv preprint arXiv:2203.04802}, 2022.

\bibitem{liang2021swinir}
Jingyun Liang, Jiezhang Cao, Guolei Sun, Kai Zhang, Luc Van~Gool, and Radu
  Timofte.
\newblock Swinir: Image restoration using swin transformer.
\newblock {\em arXiv preprint arXiv:2108.10257}, 2021.

\bibitem{barf}
Chen-Hsuan Lin, Wei-Chiu Ma, Antonio Torralba, and Simon Lucey.
\newblock Barf: Bundle-adjusting neural radiance fields.
\newblock In {\em Proceedings of the IEEE/CVF International Conference on
  Computer Vision (ICCV)}, pages 5741--5751, October 2021.

\bibitem{lin2023visionnerf}
Kai-En Lin, Lin Yen-Chen, Wei-Sheng Lai, Tsung-Yi Lin, Yi-Chang Shih, and Ravi
  Ramamoorthi.
\newblock Vision transformer for nerf-based view synthesis from a single input
  image.
\newblock In {\em WACV}, 2023.

\bibitem{treegan}
Xinyue Liu, Xiangnan Kong, Lei Liu, and Kuorong Chiang.
\newblock Treegan: syntax-aware sequence generation with generative adversarial
  networks.
\newblock In {\em 2018 IEEE International Conference on Data Mining (ICDM)},
  pages 1140--1145. IEEE, 2018.

\bibitem{lorensen1987marching}
William~E Lorensen and Harvey~E Cline.
\newblock Marching cubes: A high resolution 3d surface construction algorithm.
\newblock {\em ACM siggraph computer graphics}, 21(4):163--169, 1987.

\bibitem{gnerf}
Quan Meng, Anpei Chen, Haimin Luo, Minye Wu, Hao Su, Lan Xu, Xuming He, and
  Jingyi Yu.
\newblock Gnerf: Gan-based neural radiance field without posed camera.
\newblock In {\em Proceedings of the IEEE/CVF International Conference on
  Computer Vision (ICCV)}, pages 6351--6361, October 2021.

\bibitem{nerf}
Ben Mildenhall, Pratul~P Srinivasan, Matthew Tancik, Jonathan~T Barron, Ravi
  Ramamoorthi, and Ren Ng.
\newblock Nerf: Representing scenes as neural radiance fields for view
  synthesis.
\newblock {\em Communications of the ACM}, 65(1):99--106, 2021.

\bibitem{diffrf}
Norman M{\"u}ller, Yawar Siddiqui, Lorenzo Porzi, Samuel~Rota Bul{\`o}, Peter
  Kontschieder, and Matthias Nie{\ss}ner.
\newblock Diffrf: Rendering-guided 3d radiance field diffusion.
\newblock {\em arXiv preprint arXiv:2212.01206}, 2022.

\bibitem{autorf}
Norman M{\"{u}}ller, Andrea Simonelli, Lorenzo Porzi, Samuel~Rota Bulò,
  Matthias Nie{\ss}ner, and Peter Kontschieder.
\newblock Autorf: Learning 3d object radiance fields from single view
  observations.
\newblock In {\em Proceedings of the IEEE/CVF Conference on Computer Vision and
  Pattern Recognition (CVPR)}, June 2022.

\bibitem{mueller2022instant}
Thomas M\"uller, Alex Evans, Christoph Schied, and Alexander Keller.
\newblock Instant neural graphics primitives with a multiresolution hash
  encoding.
\newblock {\em ACM Trans. Graph.}, 41(4):102:1--102:15, July 2022.

\bibitem{hncompress1}
Phuoc Nguyen, T.~Tran, Ky~Le, Sunil Gupta, Santu Rana, Dang Nguyen, Trong
  Nguyen, Shannon Ryan, and Svetha Venkatesh.
\newblock Fast conditional network compression using bayesian hypernetworks.
\newblock In {\em ECML/PKDD}, 2022.

\bibitem{niemeyer2022regnerf}
Michael Niemeyer, Jonathan~T Barron, Ben Mildenhall, Mehdi~SM Sajjadi, Andreas
  Geiger, and Noha Radwan.
\newblock Regnerf: Regularizing neural radiance fields for view synthesis from
  sparse inputs.
\newblock In {\em Proceedings of the IEEE/CVF Conference on Computer Vision and
  Pattern Recognition}, pages 5480--5490, 2022.

\bibitem{unisurf}
Michael Oechsle, Songyou Peng, and Andreas Geiger.
\newblock Unisurf: Unifying neural implicit surfaces and radiance fields for
  multi-view reconstruction.
\newblock In {\em International Conference on Computer Vision (ICCV)}, 2021.

\bibitem{deepsdf}
Jeong~Joon Park, Peter Florence, Julian Straub, Richard Newcombe, and Steven
  Lovegrove.
\newblock Deepsdf: Learning continuous signed distance functions for shape
  representation.
\newblock In {\em Proceedings of the IEEE/CVF conference on computer vision and
  pattern recognition}, pages 165--174, 2019.

\bibitem{dreamfusion}
Ben Poole, Ajay Jain, Jonathan~T. Barron, and Ben Mildenhall.
\newblock Dreamfusion: Text-to-3d using 2d diffusion.
\newblock {\em arXiv}, 2022.

\bibitem{clip}
Alec Radford, Jong~Wook Kim, Chris Hallacy, Aditya Ramesh, Gabriel Goh,
  Sandhini Agarwal, Girish Sastry, Amanda Askell, Pamela Mishkin, Jack Clark,
  et~al.
\newblock Learning transferable visual models from natural language
  supervision.
\newblock In {\em International conference on machine learning}, pages
  8748--8763. PMLR, 2021.

\bibitem{vqvae2}
Ali Razavi, A{\"a}ron van~den Oord, and Oriol Vinyals.
\newblock Generating diverse high-fidelity images with vq-vae-2.
\newblock {\em ArXiv}, abs/1906.00446, 2019.

\bibitem{lolnerf}
Daniel Rebain, Mark Matthews, Kwang~Moo Yi, Dmitry Lagun, and Andrea
  Tagliasacchi.
\newblock Lolnerf: Learn from one look.
\newblock In {\em Proceedings of the IEEE/CVF Conference on Computer Vision and
  Pattern Recognition}, pages 1558--1567, 2022.

\bibitem{depthpriorsnerf}
Barbara Roessle, Jonathan~T. Barron, Ben Mildenhall, Pratul~P. Srinivasan, and
  Matthias Nie{\ss}ner.
\newblock Dense depth priors for neural radiance fields from sparse input
  views.
\newblock In {\em Proceedings of the IEEE/CVF Conference on Computer Vision and
  Pattern Recognition (CVPR)}, June 2022.

\bibitem{sajjadi2022scene}
Mehdi~SM Sajjadi, Henning Meyer, Etienne Pot, Urs Bergmann, Klaus Greff, Noha
  Radwan, Suhani Vora, Mario Lu{\v{c}}i{\'c}, Daniel Duckworth, Alexey
  Dosovitskiy, et~al.
\newblock Scene representation transformer: Geometry-free novel view synthesis
  through set-latent scene representations.
\newblock In {\em Proceedings of the IEEE/CVF Conference on Computer Vision and
  Pattern Recognition}, pages 6229--6238, 2022.

\bibitem{sajnani2022_condor}
Rahul Sajnani, Adrien Poulenard, Jivitesh Jain, Radhika Dua, Leonidas~J.
  Guibas, and Srinath Sridhar.
\newblock Condor: Self-supervised canonicalization of 3d pose for partial
  shapes.
\newblock In {\em The IEEE Conference on Computer Vision and Pattern
  Recognition (CVPR)}, June 2022.

\bibitem{graf}
Katja Schwarz, Yiyi Liao, Michael Niemeyer, and Andreas Geiger.
\newblock Graf: Generative radiance fields for 3d-aware image synthesis.
\newblock In {\em Advances in Neural Information Processing Systems (NeurIPS)},
  2020.

\bibitem{sen2022inrv}
Bipasha Sen, Aditya Agarwal, Vinay~P Namboodiri, and C.V. Jawahar.
\newblock {INR}-v: A continuous representation space for video-based generative
  tasks.
\newblock {\em Transactions on Machine Learning Research}, 2022.

\bibitem{scarp}
Bipasha Sen, Aditya Agarwal, Gaurav Singh, Brojeshwar B., Srinath Sridhar, and
  Madhava Krishna.
\newblock Scarp: 3d shape completion in arbitrary poses for improved grasping.
\newblock In {\em 2023 IEEE International Conference on Robotics and Automation
  (ICRA)}, pages 3838--3845, 2023.

\bibitem{hnfewshot1}
Marcin Sendera, Marcin Przewiezlikowski, Konrad Karanowski, Maciej Zieba, Jacek
  Tabor, and Przemysław Spurek.
\newblock Hypershot: Few-shot learning by kernel hypernetworks.
\newblock {\em 2023 IEEE/CVF Winter Conference on Applications of Computer
  Vision (WACV)}, pages 2468--2477, 2022.

\bibitem{rndfs}
Anthony Simeonov, Yilun Du, Yen-Chen Lin, Alberto~Rodriguez Garcia, Leslie~Pack
  Kaelbling, Tom{\'a}s Lozano-P{\'e}rez, and Pulkit Agrawal.
\newblock {SE}(3)-equivariant relational rearrangement with neural descriptor
  fields.
\newblock In {\em 6th Annual Conference on Robot Learning}, 2022.

\bibitem{ndfs}
Anthony Simeonov, Yilun Du, Andrea Tagliasacchi, Joshua~B. Tenenbaum, Alberto
  Rodriguez, Pulkit Agrawal, and Vincent Sitzmann.
\newblock Neural descriptor fields: Se(3)-equivariant object representations
  for manipulation.
\newblock 2022.

\bibitem{sparsepose}
Samarth Sinha, Jason~Y Zhang, Andrea Tagliasacchi, Igor Gilitschenski, and
  David~B Lindell.
\newblock Sparsepose: Sparse-view camera pose regression and refinement.
\newblock {\em arXiv preprint arXiv:2211.16991}, 2022.

\bibitem{siren}
Vincent Sitzmann, Julien Martel, Alexander Bergman, David Lindell, and Gordon
  Wetzstein.
\newblock Implicit neural representations with periodic activation functions.
\newblock {\em Advances in Neural Information Processing Systems},
  33:7462--7473, 2020.

\bibitem{lfns}
Vincent Sitzmann, Semon Rezchikov, Bill Freeman, Josh Tenenbaum, and Fredo
  Durand.
\newblock Light field networks: Neural scene representations with
  single-evaluation rendering.
\newblock {\em Advances in Neural Information Processing Systems},
  34:19313--19325, 2021.

\bibitem{3dhypernet2}
Vincent Sitzmann, Michael Zollhoefer, and Gordon Wetzstein.
\newblock Scene representation networks: Continuous 3d-structure-aware neural
  scene representations.
\newblock In H.~Wallach, H.~Larochelle, A.~Beygelzimer, F.~d\textquotesingle
  Alch\'{e}-Buc, E.~Fox, and R.~Garnett, editors, {\em Advances in Neural
  Information Processing Systems}, volume~32. Curran Associates, Inc., 2019.

\bibitem{inr-gan}
Ivan Skorokhodov, Savva Ignatyev, and Mohamed Elhoseiny.
\newblock Adversarial generation of continuous images.
\newblock In {\em Proceedings of the IEEE/CVF Conference on Computer Vision and
  Pattern Recognition}, pages 10753--10764, 2021.

\bibitem{stylegan-v}
Ivan Skorokhodov, Sergey Tulyakov, and Mohamed Elhoseiny.
\newblock Stylegan-v: A continuous video generator with the price, image
  quality and perks of stylegan2.
\newblock In {\em Proceedings of the IEEE/CVF Conference on Computer Vision and
  Pattern Recognition}, pages 3626--3636, 2022.

\bibitem{sun2022direct}
Cheng Sun, Min Sun, and Hwann-Tzong Chen.
\newblock Direct voxel grid optimization: Super-fast convergence for radiance
  fields reconstruction.
\newblock In {\em Proceedings of the IEEE/CVF Conference on Computer Vision and
  Pattern Recognition}, pages 5459--5469, 2022.

\bibitem{tancik2022block}
Matthew Tancik, Vincent Casser, Xinchen Yan, Sabeek Pradhan, Ben Mildenhall,
  Pratul~P Srinivasan, Jonathan~T Barron, and Henrik Kretzschmar.
\newblock Block-nerf: Scalable large scene neural view synthesis.
\newblock In {\em Proceedings of the IEEE/CVF Conference on Computer Vision and
  Pattern Recognition}, pages 8248--8258, 2022.

\bibitem{learnedinit}
Matthew Tancik, Ben Mildenhall, Terrance Wang, Divi Schmidt, Pratul~P.
  Srinivasan, Jonathan~T. Barron, and Ren Ng.
\newblock Learned initializations for optimizing coordinate-based neural
  representations.
\newblock In {\em CVPR}, 2021.

\bibitem{tewari2022advances}
Ayush Tewari, Justus Thies, Ben Mildenhall, Pratul Srinivasan, Edgar Tretschk,
  Wang Yifan, Christoph Lassner, Vincent Sitzmann, Ricardo Martin-Brualla,
  Stephen Lombardi, et~al.
\newblock Advances in neural rendering.
\newblock In {\em Computer Graphics Forum}, volume~41, pages 703--735. Wiley
  Online Library, 2022.

\bibitem{continualhypernet}
Johannes Von~Oswald, Christian Henning, Jo{\~a}o Sacramento, and Benjamin~F
  Grewe.
\newblock Continual learning with hypernetworks.
\newblock {\em arXiv preprint arXiv:1906.00695}, 2019.

\bibitem{clipnerf}
Can Wang, Menglei Chai, Mingming He, Dongdong Chen, and Jing Liao.
\newblock Clip-nerf: Text-and-image driven manipulation of neural radiance
  fields.
\newblock In {\em Proceedings of the IEEE/CVF Conference on Computer Vision and
  Pattern Recognition}, pages 3835--3844, 2022.

\bibitem{nocs}
He~Wang, Srinath Sridhar, Jingwei Huang, Julien Valentin, Shuran Song, and
  Leonidas~J. Guibas.
\newblock Normalized object coordinate space for category-level 6d object pose
  and size estimation.
\newblock In {\em The IEEE Conference on Computer Vision and Pattern
  Recognition (CVPR)}, June 2019.

\bibitem{neus}
Peng Wang, Lingjie Liu, Yuan Liu, Christian Theobalt, Taku Komura, and Wenping
  Wang.
\newblock Neus: Learning neural implicit surfaces by volume rendering for
  multi-view reconstruction.
\newblock {\em NeurIPS}, 2021.

\bibitem{nerfminus}
Zirui Wang, Shangzhe Wu, Weidi Xie, Min Chen, and Victor~Adrian Prisacariu.
\newblock Ne{RF}$--$: Neural radiance fields without known camera parameters.
\newblock {\em arXiv preprint arXiv:2102.07064}, 2021.

\bibitem{neuralfield3Dreconstruction}
Francis Williams, Zan Gojcic, Sameh Khamis, Denis Zorin, Joan Bruna, Sanja
  Fidler, and Or~Litany.
\newblock Neural fields as learnable kernels for 3d reconstruction.
\newblock In {\em Proceedings of the IEEE/CVF Conference on Computer Vision and
  Pattern Recognition}, pages 18500--18510, 2022.

\bibitem{xiangli2021citynerf}
Yuanbo Xiangli, Linning Xu, Xingang Pan, Nanxuan Zhao, Anyi Rao, Christian
  Theobalt, Bo~Dai, and Dahua Lin.
\newblock Citynerf: Building nerf at city scale.
\newblock {\em arXiv preprint arXiv:2112.05504}, 2021.

\bibitem{xie2022_neuralfields}
Yiheng Xie, Towaki Takikawa, Shunsuke Saito, Or~Litany, Shiqin Yan, Numair
  Khan, Federico Tombari, James Tompkin, Vincent Sitzmann, and Srinath Sridhar.
\newblock Neural fields in visual computing and beyond.
\newblock {\em Computer Graphics Forum}, 2022.

\bibitem{xu2022sinnerf}
Dejia Xu, Yifan Jiang, Peihao Wang, Zhiwen Fan, Humphrey Shi, and Zhangyang
  Wang.
\newblock Sinnerf: Training neural radiance fields on complex scenes from a
  single image.
\newblock In {\em Computer Vision--ECCV 2022: 17th European Conference, Tel
  Aviv, Israel, October 23--27, 2022, Proceedings, Part XXII}, pages 736--753.
  Springer, 2022.

\bibitem{volsdf}
Lior Yariv, Jiatao Gu, Yoni Kasten, and Yaron Lipman.
\newblock Volume rendering of neural implicit surfaces.
\newblock In {\em Thirty-Fifth Conference on Neural Information Processing
  Systems}, 2021.

\bibitem{yu2021plenoxels}
Alex Yu, Sara Fridovich-Keil, Matthew Tancik, Qinhong Chen, Benjamin Recht, and
  Angjoo Kanazawa.
\newblock Plenoxels: Radiance fields without neural networks.
\newblock {\em arXiv preprint arXiv:2112.05131}, 2021.

\bibitem{yu2021plenoctrees}
Alex Yu, Ruilong Li, Matthew Tancik, Hao Li, Ren Ng, and Angjoo Kanazawa.
\newblock Plenoctrees for real-time rendering of neural radiance fields.
\newblock In {\em Proceedings of the IEEE/CVF International Conference on
  Computer Vision}, pages 5752--5761, 2021.

\bibitem{pixelnerf}
Alex Yu, Vickie Ye, Matthew Tancik, and Angjoo Kanazawa.
\newblock {pixelNeRF}: Neural radiance fields from one or few images.
\newblock In {\em CVPR}, 2021.

\bibitem{digan}
Sihyun Yu, Jihoon Tack, Sangwoo Mo, Hyunsu Kim, Junho Kim, Jung-Woo Ha, and
  Jinwoo Shin.
\newblock Generating videos with dynamics-aware implicit generative adversarial
  networks.
\newblock {\em ArXiv}, abs/2202.10571, 2022.

\bibitem{neuralfield3Dshapegen}
Biao Zhang, Jiapeng Tang, Matthias Niessner, and Peter Wonka.
\newblock 3dshape2vecset: A 3d shape representation for neural fields and
  generative diffusion models.
\newblock {\em arXiv preprint arXiv:2301.11445}, 2023.

\bibitem{lpips}
Richard Zhang, Phillip Isola, Alexei~A Efros, Eli Shechtman, and Oliver Wang.
\newblock The unreasonable effectiveness of deep features as a perceptual
  metric.
\newblock In {\em CVPR}, 2018.

\bibitem{zhou2021CIPS3D}
Peng Zhou, Lingxi Xie, Bingbing Ni, and Qi~Tian.
\newblock {{CIPS}}-{{3D}}: A {{3D}}-{{Aware Generator}} of {{GANs Based}} on
  {{Conditionally}}-{{Independent Pixel Synthesis}}.
\newblock 2021.

\end{thebibliography}
\bibliographystyle{plain}

\newpage
\appendix

\section{Experiments}
\vspace{-10px}

\begin{table}[h]
    \centering
    \adjustbox{max width=0.85\linewidth}{
    \begin{tabular}
    {r|l|ccc|ccc}
    \toprule
    & & \multicolumn{3}{c|}{Chairs} & \multicolumn{3}{c}{Sofa} \\
        & & PSNR$\uparrow$ & SSIM$\uparrow$ & LPIPS$\downarrow$ 
        & PSNR$\uparrow$ & SSIM$\uparrow$ & LPIPS$\downarrow$\\
    \midrule
    \multirow{2}{*}{ABO} & PixelNeRF~\cite{pixelnerf} & 19.00 & 0.74 & 0.343 & 18.49 & 0.77 & 0.351 \\
    & CodeNeRF~\cite{jang2021codenerf} & 20.51 & 0.75 & 0.264 & 20.38 & 0.77 & 0.31\\
    & \acro (Ours) & \textbf{25.92} & \textbf{0.91} & \textbf{0.093} & \textbf{26.73} & \textbf{0.91} & \textbf{0.098}\\
    & \;\;\;\;\;\;w/o denoising & 24.83 & 0.87 & 0.12& 25.68 & 0.87 & 0.14\\
    \bottomrule
    \end{tabular}}
    \vspace{0.2cm}
    \caption{\textbf{Generalization}. Comparison of single-view NeRF generation on the ABO dataset. Metrics are computed on renderings of resolution 128 × 128. \acro significantly outperforms PixelNeRF~\cite{pixelnerf} and CodeNeRF~\cite{jang2021codenerf} on all of the metrics in both object categories. }
    \label{tab:abosupp_gen_exps}
    \vspace{-1em}
\end{table}

\subsection{Additional Architectural Details}

We provide the network architecture in the main paper, Section 4. During training, we use Adam Optimizer, with a learning rate of $1e-3$ with $\beta_1=0.9$ and $\beta_2=0.99$, along with a lambda LR scheduler\footnote{\textcolor{blue}{\href{https://pytorch.org/docs/stable/generated/torch.optim.lr_scheduler.LambdaLR.html}{https://pytorch.org/docs/stable/generated/torch.optim.lr\_scheduler.LambdaLR.html}}}. We use the PyTorch implementation InstantNGP\footnote{\textcolor{blue}{\href{https://github.com/ashawkey/torch-ngp}{https://github.com/ashawkey/torch-ngp}}} and provide the training, inference, and metric computation code in the Appendix. 

\subsection{Metrics and Additional Comparisons}
\label{sec:suppmetric}


To the best of our knowledge, we are the first work to perform single-view NeRF generation at a resolution of 512$\times$512. Therefore, we set a benchmark on the ABO dataset against PixelNeRF in the main paper, Table 1. 
However, it is worth noting that PixelNeRF was originally trained at a resolution of 128$\times$128. Therefore, we also compare with PixelNeRF on ABO at a resolution of 128$\times$128 in 
the Appendix, \Cref{tab:abosupp_gen_exps} and  
\Cref{fig:qualabo128}. To do so, we retrain PixelNeRF on 128$\times$128 by downsampling the ground truth datapoints in ABO. 
Further, we show qualitative results in the main paper on SRN~\cite{3dhypernet2} against VisionNeRF~\cite{lin2023visionnerf}, FE-NVS~\cite{guo2022fast}, CodeNeRF~\cite{jang2021codenerf}, and PixelNeRF at 128$\times$128. In the Appendix, we show quantitative results in \Cref{tab:srn_gen_exps}. Lastly, in \Cref{tab:denoise}, we show additional ablation on Denoise and Finetune (see main paper, Section 3.1-Step 2). In this table, we primarily evaluate the geometric consistency before and after the denoising step. As explained in the main paper, Section 4, we use Chamfer's Distance (CD$\downarrow$) to compute the geometric consistency. 

To compute CD, we first train the ground truth NeRFs by optimizing InstantNGP~\cite{mueller2022instant} on the multiview ground truths. Next, we render meshes from InstantNGP's and \acro's predicted NeRF using torch-ngp's save\_mesh() implementation\footnote{\textcolor{blue}{\href{https://github.com/ashawkey/torch-ngp/blob/main/nerf/utils.py}{https://github.com/ashawkey/torch-ngp/blob/main/nerf/utils.py}}}. From the rendered mesh, we sample 4096 points uniformly and compute CD between both the pointclouds. \textcolor{orange}{\textbf{We encourage the readers to view the supplementary video for the best experience of the qualitative results. }}

\begin{table}[t]
    \centering
    \adjustbox{max width=0.75\linewidth}{
    \begin{tabular}
    {r|l|ccc|ccc}
    \toprule
    & & \multicolumn{3}{c|}{Chairs} & \multicolumn{3}{c}{Cars} \\
        & & PSNR$\uparrow$ & SSIM$\uparrow$ & LPIPS$\downarrow$ 
        & PSNR$\uparrow$ & SSIM$\uparrow$ & LPIPS$\downarrow$ \\
    \midrule
    \multirow{6}{*}{SRN} & PixelNeRF~\cite{pixelnerf} & 23.72 & 0.90 & 0.128 
    & 23.17 & 0.89 & 0.146 \\
    & CodeNeRF~\cite{jang2021codenerf} & 23.39 & 0.87 & 0.166 & 
    22.73 & 0.89 & 0.128 \\
    & FE-NVS~\cite{guo2022fast} & 23.21 &\textbf{ 0.92} & \textbf{0.077} 
    & 22.83 & \textbf{0.91} & 0.099 \\ 
    & VisionNeRF~\cite{lin2023visionnerf} & \textbf{24.48} & \textbf{0.92} & \textbf{0.077} 
    & 22.88 & 0.90 & \textbf{0.084} \\
    & \acro & 22.80 & 0.88 & 0.13 & \textbf{23.48} & \textbf{0.91} & 0.09 \\
    & \;\;\;\;\;\;w/o Denoise & 21.02 & 0.87 & 0.14 
    & 21.30 & 0.88 & 0.111 \\
    \bottomrule
    \end{tabular}}
    \vspace{0.1cm}
    \caption{\textbf{Generalization}. Comparison of single-view NeRF generation on the SRN dataset. Metrics are computed on renderings of resolution 128 × 128. Results of all the models (except \acro) are taken from the official papers. \acro performs comparably with the existing baselines. Note, we do not incorporate the second step of \acro, Denoise and Finetune (main paper, Section 3.1). }
    \label{tab:srn_gen_exps}
    \vspace{-1em}
\end{table}

\begin{figure}[t]
    \centering
    \includegraphics[width=\linewidth]{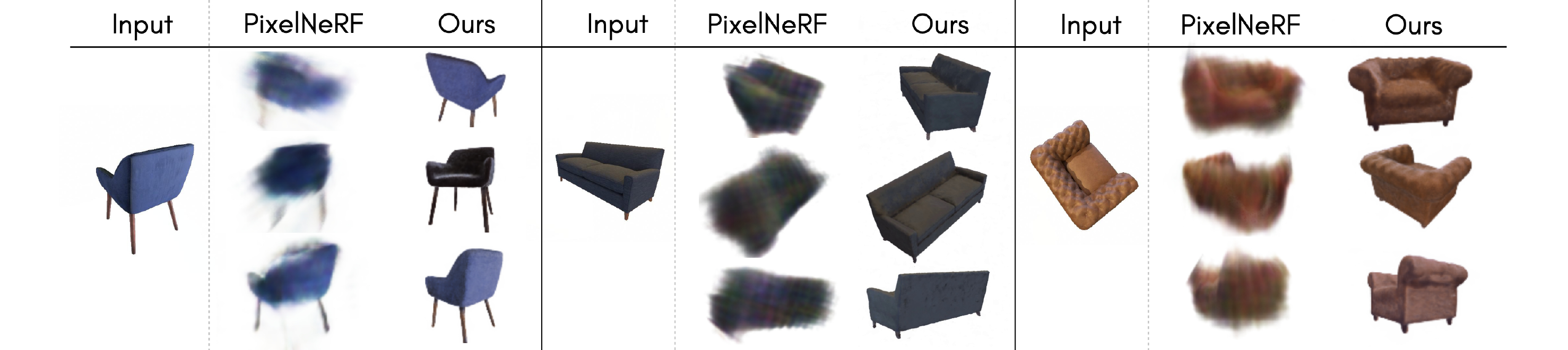}
    \caption{\textbf{Qualitative results of single-view NeRF generation} on ABO dataset at a resolution of 128$\times$128. HyP-NeRF preserves fine details even at this low resolution. PixelNeRF improves in quality when compared to 512$\times$512 (see main paper, Figure 4). However, it still struggles to model the fine texture and shape details in the ABO dataset and performs subpar to \acro.}
    \label{fig:qualabo128}
    \vspace{1em}
    \includegraphics[width=\linewidth]{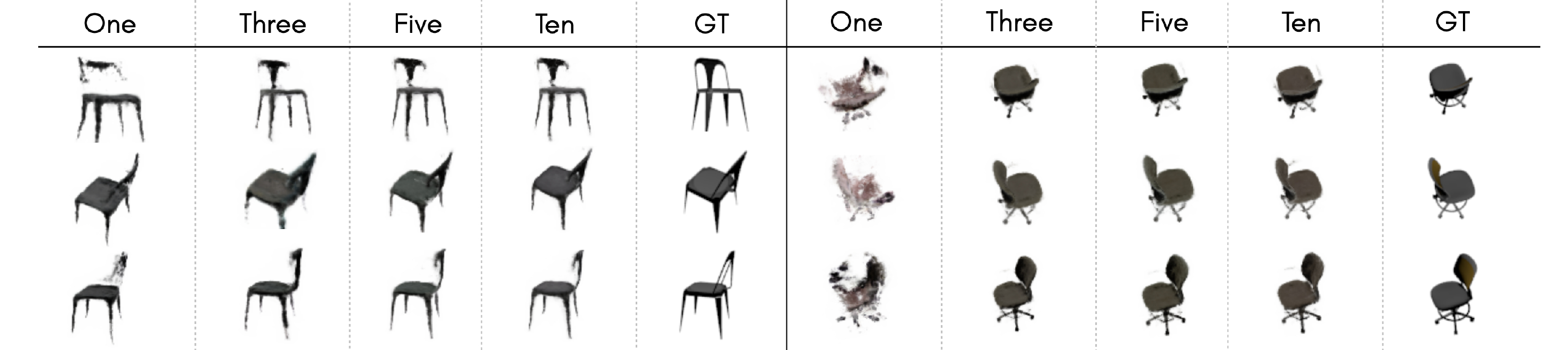}
    \caption{\small \textbf{Multiview test-time optimization on SRN Chairs}. \acro can perform test-time optimization (see main paper, Section 3.1) with any number of views. In this qualitative result, we start with an instance that did not optimize well through a single pose (because of a challenging viewpoint) and show the improvement in quality of the generated NeRF as we increase the number of views (ie. coverage) for optimization. The header indicate the number of views used for the optimization. As shown, the difference in quality between five and ten poses is insignificant. The rightmost result shows drastic improvement in the render quality from one to three views showcasing the impact of pose on test-time optimization. }
    \label{fig:srnmultiview}
    \vspace{-1em}
\end{figure}



\subsection{Generalization} 
To ensure that \acro can model novel NeRF instances unseen at the time of training, we rely on the conditional task of ``single-view NeRF generation". In the main paper, we show experiments on ABO dataset at 512$\times$512 resolution; in the Appendix, we make comparisons on a lower resolution of 128$\times$128 on ABO against PixelNeRF in \Cref{tab:abosupp_gen_exps} and on SRN against the existing baselines in \Cref{tab:srn_gen_exps}. As can be seen, we significantly outperform PixelNeRF on the ABO dataset and perform comparably with the existing baselines on the SRN dataset. 
\underline{\textbf{Note}} that we do not employ the Denoise and Finetune step (see main paper, Section 3.1) in SRN. 
However, another reason for our low performance on SRN (when compared to ABO) is the difference in the views adopted in ABO and SRN. 
ABO renders the 3D structure from 91 viewpoints on the upper icosphere with varying azimuth and elevation~\cite{abo}. SRN, on the other hand, renders the upper along with the lower icosphere. This includes viewpoints from the absolute bottom and top parts of the object providing insufficient context for test-time optimization. 

We observe that, in practice, our output quality improves significantly as we increase the number of viewpoints on SRN as shown qualitatively in \Cref{fig:srnmultiview}. This indicates that although \acro has modeled this particular NeRF instance, it is hard to find the NeRF through single-view optimization suggesting the need for a more robust mechanism to map to \acro's prior. 

It is also worth noting that, PixelNeRF and VisionNeRF (trained on 16 NVIDIA A100 for 5 days) are designed specifically for the task single-view NeRF generation. Whereas, we aim to train a prior and use this conditional task to validate that our learned prior can model novel instances unseen at the time of training. Further, as can be seen in \Cref{tab:abosupp_gen_exps}, \acro significantly outperforms PixelNeRF on the challenging ABO dataset with high-fidelity structure and texture, indicating that \acro is capable of modeling datasets made of fine textures and shapes as found in the real-world, that the existing work (PixelNeRF) struggle to train on. 


\begin{figure}
\hspace{-2.7em}
  \begin{subfigure}[t]{.6\textwidth}
    \centering
    \includegraphics[width=\linewidth]{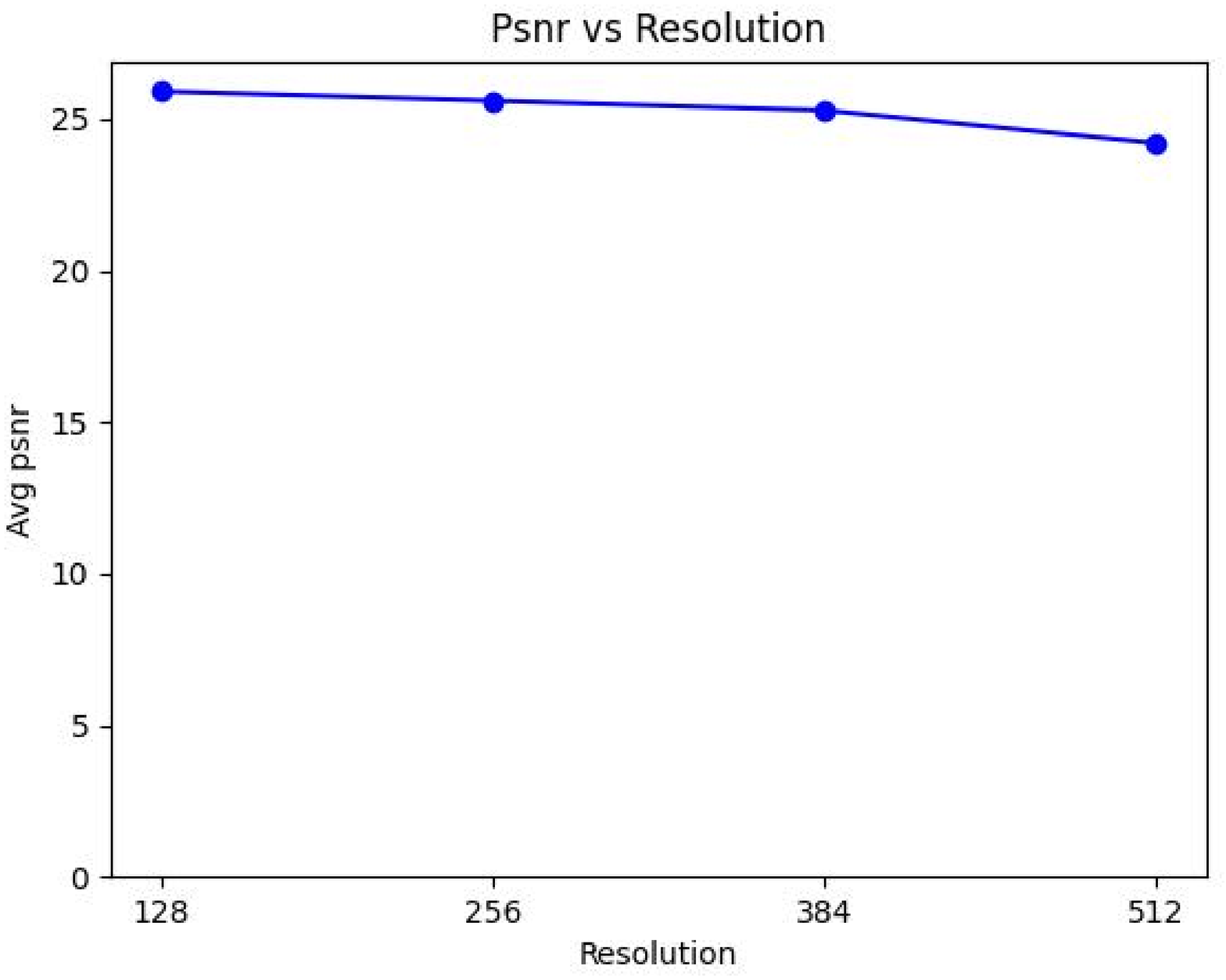}
  \end{subfigure}
  \hspace{-2.7em}
  \begin{subfigure}[t]{.6\textwidth}
    \centering
    \includegraphics[width=\linewidth]{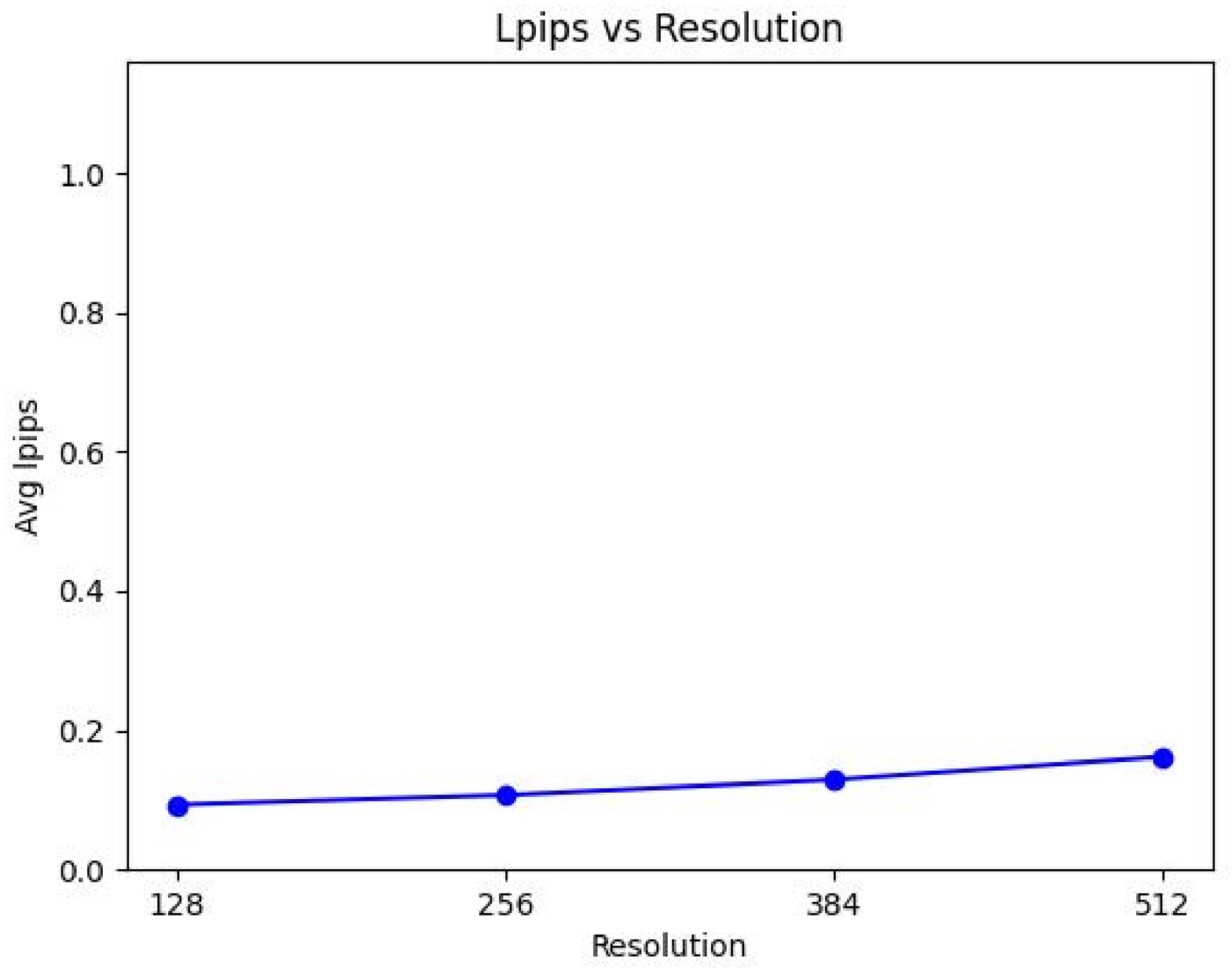}
  \end{subfigure}
\vspace{-1em}
\hspace{-2.7em}
  \begin{subfigure}[t]{.6\textwidth}
    \centering
    \includegraphics[width=\linewidth]{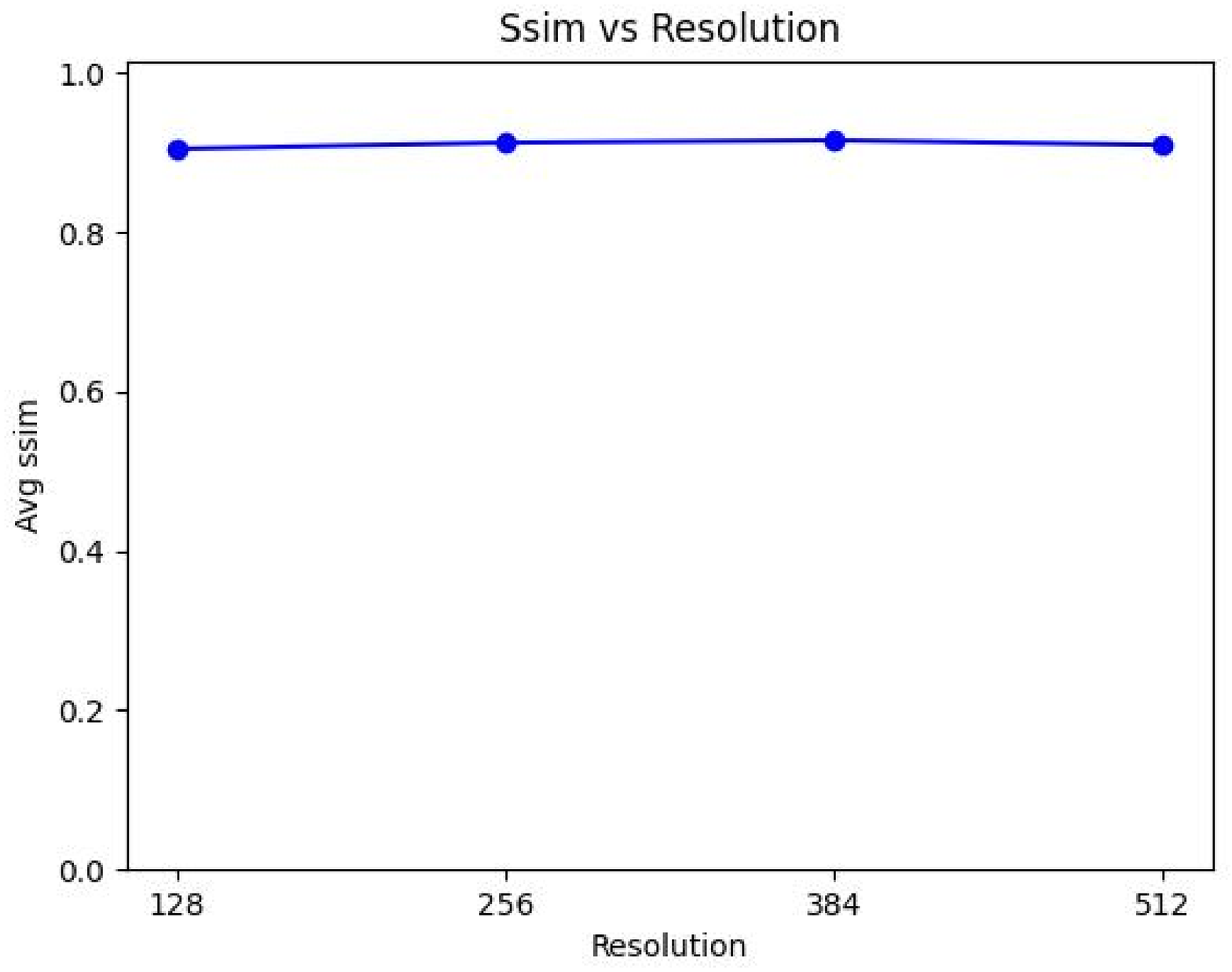}
  \end{subfigure}
  \hspace{-2.7em}
  \begin{subfigure}[t]{.6\textwidth}
    \centering
    \includegraphics[width=\linewidth]{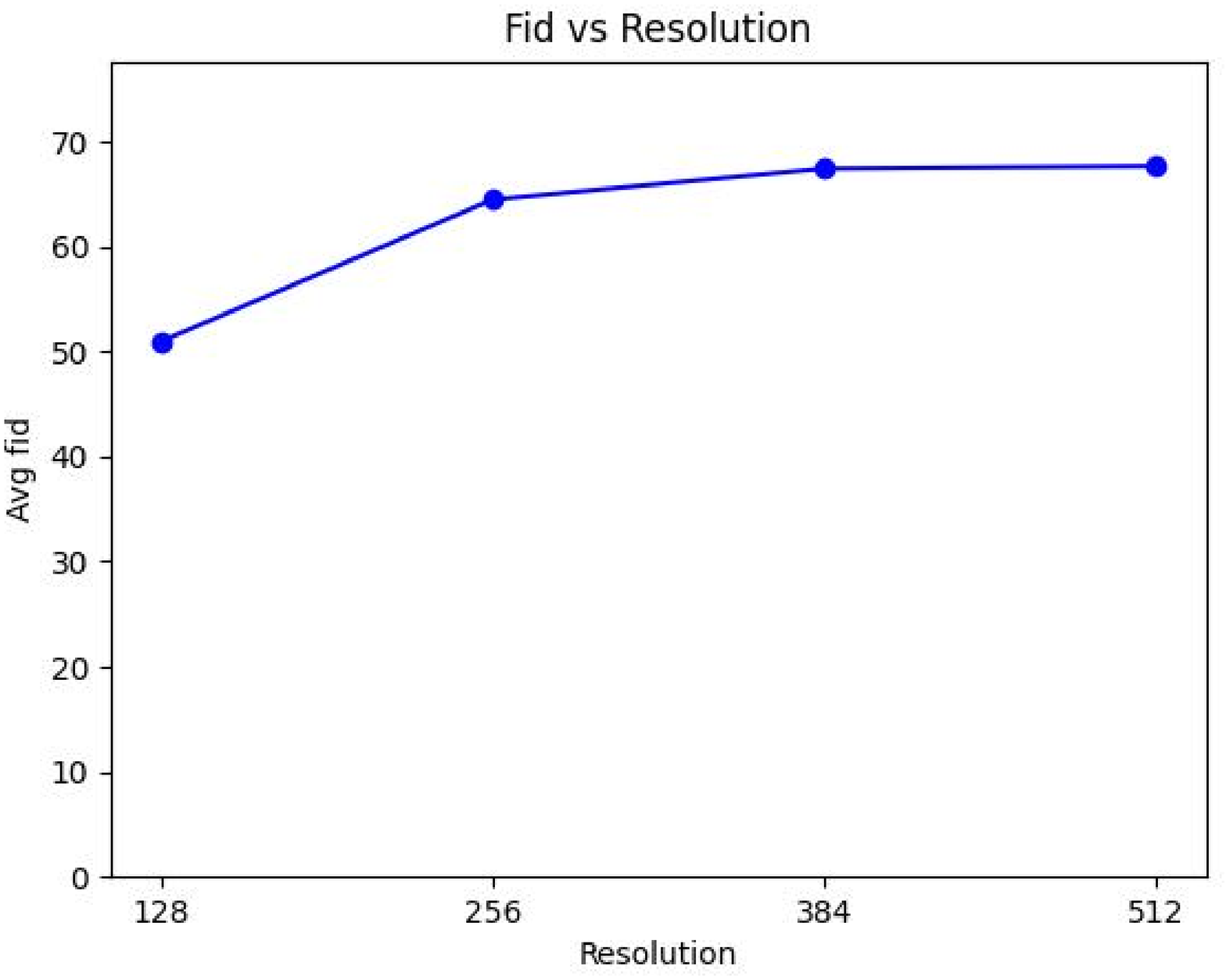}
  \end{subfigure}
  \caption{\small \textbf{Performance of \acro on multiple resolutions}. As \acro operates directly in the NeRF space, it can render the NeRFs in potentially any resolution. In this plot, we showcase \acro's performance rendered in different resolutions. As can be seen, the quality does not degrade with the resolution and \acro performs well consistently. }
  \label{fig:resolutionvsmetric}
  \vspace{-1em}
\end{figure}

\section{Additional Ablations}

\begin{wraptable}[12]{R}{4cm}
    \vspace{-1.6em}
    \centering
    \adjustbox{max width=\linewidth}{
    \begin{tabular}
    {r|c}
    \toprule
    & Chairs \\
    \acro & CD $\downarrow$ \\
    \midrule
    with D\&F & 0.0062\\
    without D\&F & 0.0064 \\
    \bottomrule
    \end{tabular}}
    \vspace{0.1cm}
    \caption{\small \textbf{Denoise and Finetune (D\&F) ablation (see main paper (Section 3.1 Step 2)}. We evaluate the geometric consistency using CD$\downarrow$ metric defined in \Cref{sec:suppmetric}.}
    \label{tab:denoise}
\end{wraptable}


\textbf{Impact of Resolution on the Quality:} As we operate directly in the NeRF space, we can essentially render the NeRFs in any resolution. In this ablation, we measure the quality of our renderings at different resolutions as shown in \Cref{fig:resolutionvsmetric}. To generate the ground truth, we perform interarea downsampling on the ABO datapoints. As expected, \acro generates high-quality NeRF in each resolution, and the quality does not degrade with the rendered resolution. As ABO consists of rendering at a resolution of 512$\times$512, we only make comparisons on the lower resolutions as bicubic upsampling on the ground truth would reduce the quality of the ground truth itself. However, to showcase our quality on higher resolution, we present our rendering at 1024$\times$1024 in the supplementary video, timestamp 01:17. 

\vspace{-5px}
\textbf{Geometric Consistency on Denoising:} As explained in the main paper, Section 3.1, we perform Denoise and Finetune by first projecting the NeRF into predefined multiview images, followed by performing image-level denoising frame-by-frame. As we only \underline{finetune} an already multiview and geometrically consistent NeRF, we observed that the finetuning is robust to minor denoised image-level multiview inconsistencies. We showcase this qualitatively in the supplementary video (timestamps 3:05-3:30), in the main paper-Figure 4, and in the Appendix-\Cref{fig:inv_1} and \Cref{fig:inv_2}. In this section, we quantitatively evaluate the geometric consistency using the CD metric as defined in \Cref{sec:suppmetric}. The results are presented in \Cref{tab:denoise}, and as expected, there is no degradation in the quality between \acro's output before and after the Denoise and Finetune process, clearly showcasing that the geometric consistency is not affected even though we only rely on a simple frame-by-frame denoising. 

\vspace{-5px}
\section{Qualitative Results}

\begin{figure}[h!]
    \centering
    \includegraphics[width=0.7\linewidth]{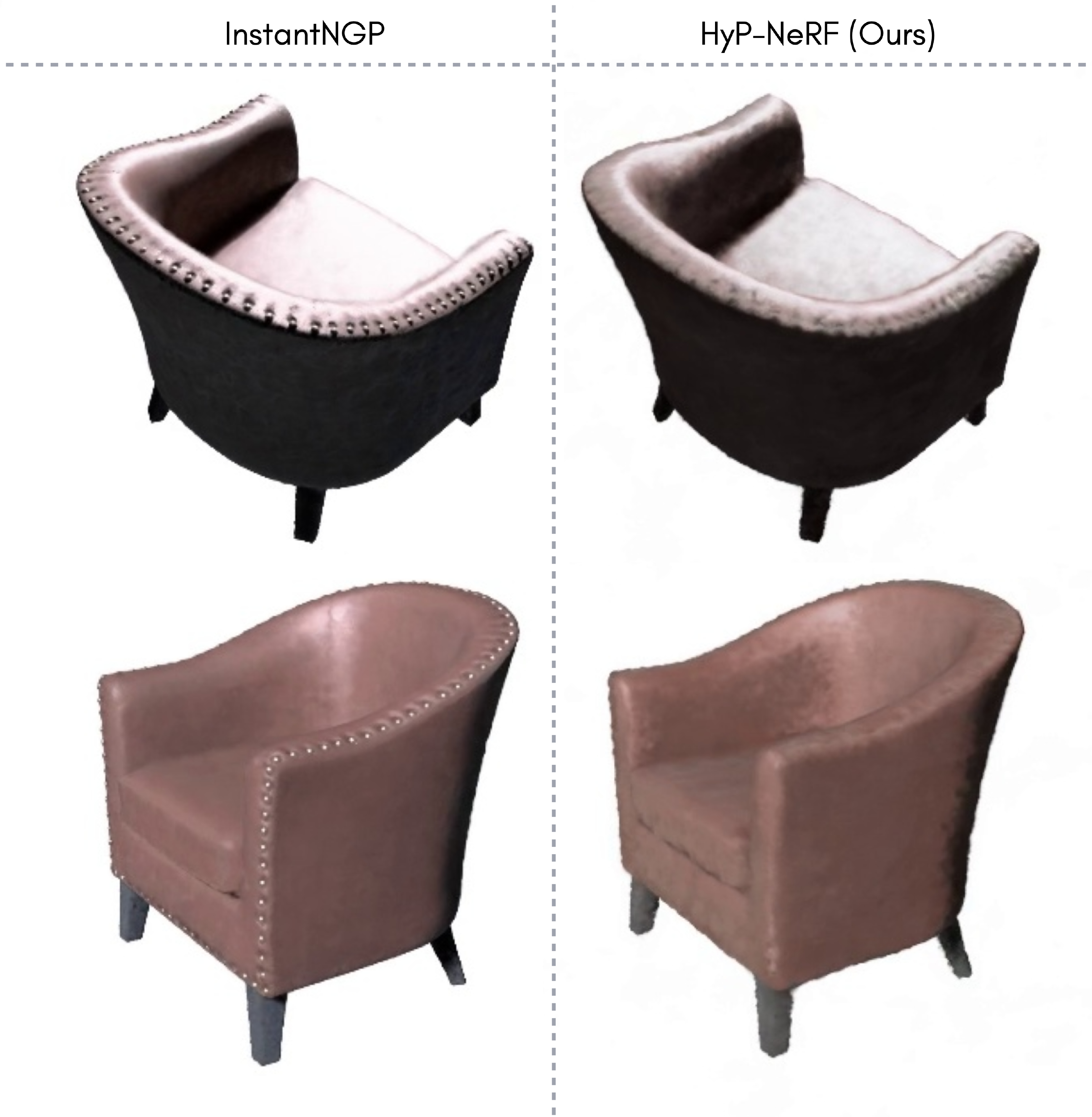}
    \caption{\textbf{Qualitative comparison on Compression}. We compare against InstantNGP~\cite{mueller2022instant}, which is trained for a specific instance. On the other hand, \acro is trained on thousands of NeRF instances. Despite that, \acro has learned to generate the NeRFs and essentially compress them almost losslessly. See the main paper, Section 4.2, for more details.}
    \label{fig:comp}
\end{figure}

\begin{figure}[h!]
    \centering
    \includegraphics[width=0.87\linewidth]{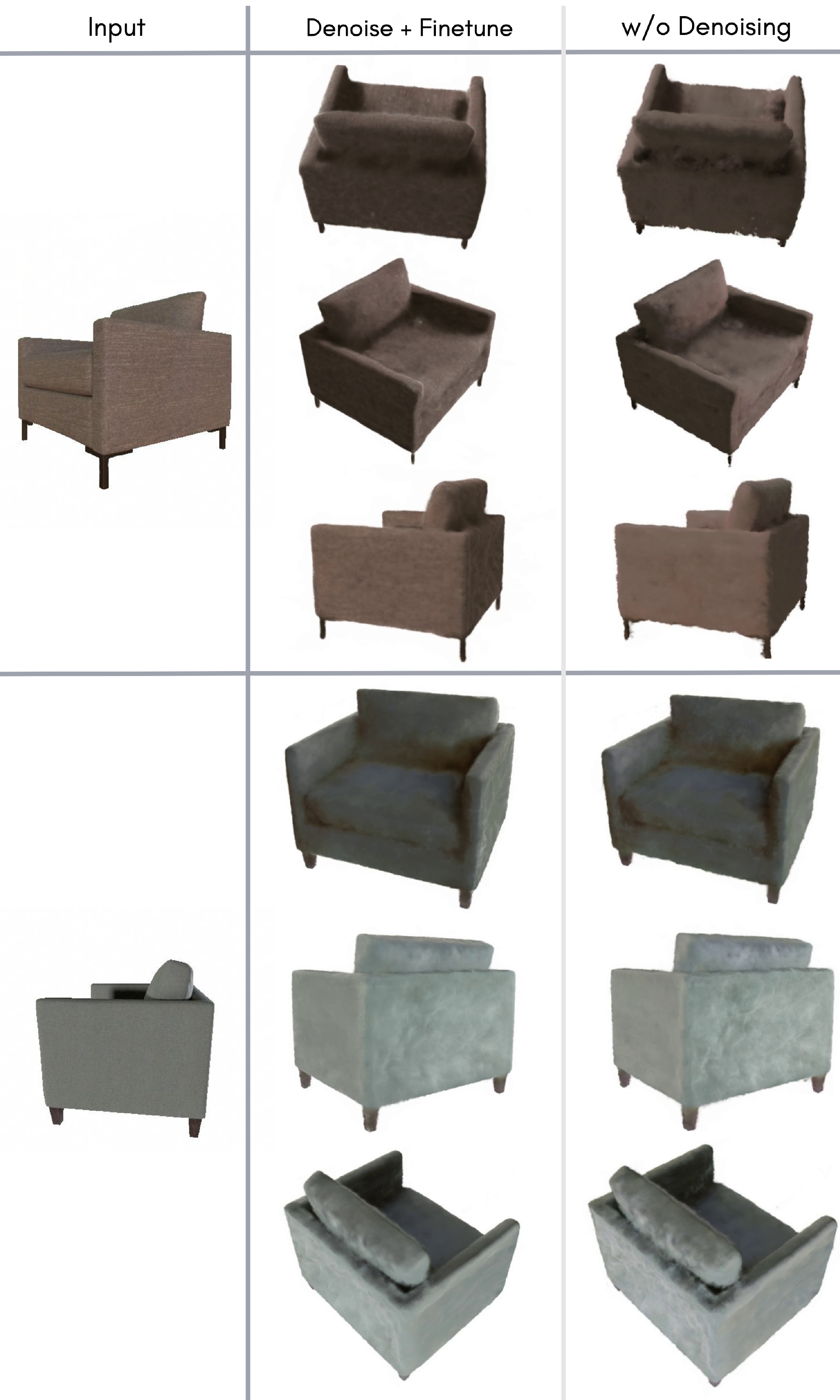}
    \caption{\textbf{Qualitative Results on Generalization}. We perform test-test optimization (see the main paper, Section 3.2) to generate NeRFs from a single input view. Our Denoise and Finetune step (see the main paper, Section 3.1) significantly improves the texture and the edges by making it smooth and even. }
    \label{fig:inv_1}
\end{figure}

\begin{figure}[h!]
    \centering
    \includegraphics[width=0.87\linewidth]{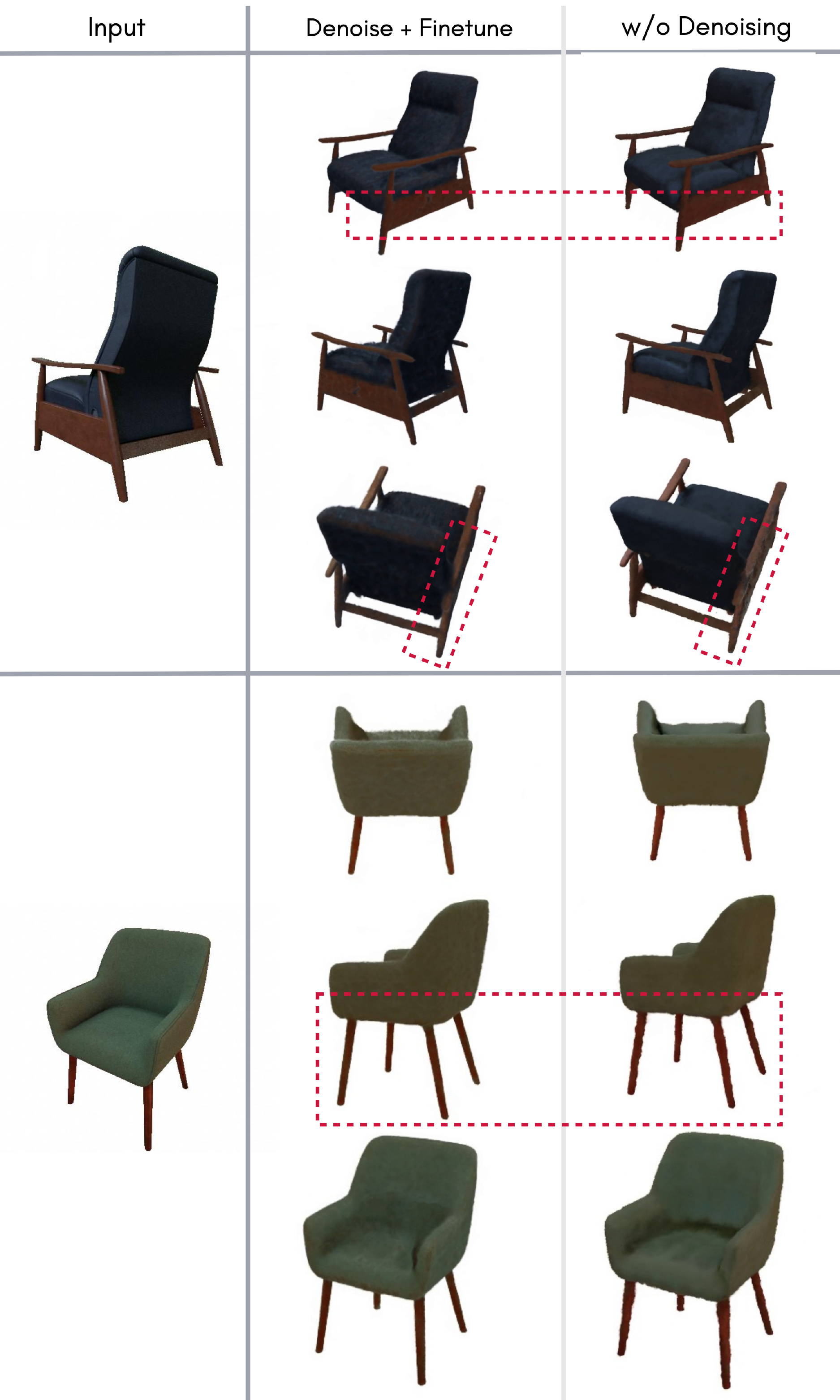}
    \caption{\textbf{Qualitative Results on Generalization}. We perform test-test optimization (see the main paper, Section 3.2) to generate NeRFs from a single input view. Denoise and Finetune (see the main paper, Section 3.1) improves the quality of the outputs, for example, the legs are clearly more evened out and noiseless in the bottom example. The difference is, however, less drastic in the top example.}
    \label{fig:inv_2}
\end{figure}

In this section, we show the qualitative results in higher resolution. \Cref{fig:comp} presents the comparison between InstantNGP, trained on a single instance, against \acro trained on thousands of NeRF instances, thereby compressing the instances to a single network (see the main paper, Section 4.2). \Cref{fig:inv_1} and \Cref{fig:inv_2} present qualitative results on inversion and highlight the difference before and after the Denoise and Finetune step (see main paper, Section 3.1). 


\end{document}